\documentclass{article}

\PassOptionsToPackage{numbers}{natbib}
\usepackage[dblblindworkshop, final]{neurips_2026}
\workshoptitle{Workshop on Linguistic Principles for Foundation Models (LP4FM)}

\usepackage[utf8]{inputenc} 
\usepackage[T1]{fontenc}    
\usepackage{hyperref}       
\usepackage{url}            
\usepackage{booktabs}       
\usepackage{amsfonts}       
\usepackage{nicefrac}       
\usepackage{microtype}      
\usepackage{xcolor}         
\usepackage{comment}
\usepackage{graphicx}
\usepackage{subcaption}
\usepackage[export]{adjustbox}
\usepackage{amsmath}
\usepackage{booktabs}
\usepackage{multirow}
\usepackage{wrapfig}
\usepackage{placeins}

\title{The Multilingual Artificial Hivemind:\\
Do All LLMs Think Alike in Every Language?}
\workshoptitle{Workshop on Linguistic Principles for Foundation Models (LP4FM)}

\author{%
  Xaver Raphael Davey \\
  KTH Royal Institute of Technology \\
  \texttt{xrdavey@kth.se} \\
  \And
  Shubhra Mishra \\
  KTH Royal Institute of Technology \\
  \texttt{shubhram@kth.se} \\
  \And
  Anton Adel\"ow \\
  KTH Royal Institute of Technology \\
  \texttt{aadelow@kth.se} \\
  \And
  Siyao Liu \\
  KTH Royal Institute of Technology \\
  \texttt{siyaol@kth.se} \\
}

\begin{document}

\maketitle

\begin{abstract}
Text generated by Large Language Models (LLMs) often remains limited in diversity compared to that produced by humans. Recent advances demonstrate the presence of inter- and intra-model collapse in open-ended generation of English text, raising concerns about the long-term homogenization of LLM, and more importantly, human thought. However, our understanding of whether this homogenization extends to languages beyond English is limited. In this work, we investigate the role of language on output diversity. Specifically, we aim to understand whether multilingual LLM-generated text is also homogeneous, and whether this homogeneity matches what we observe in humans. Across 23 open-weight models and 13 languages, we find that the artificial hivemind effect is pervasive but uneven in its extent. The intra-model effect holds in three different embedding spaces and under a lexical metric that uses no embedding model, so it is not an artifact of how diversity is measured. The size of the homogenization gap is associated with a model's competence in a language, measured on a held-out translation benchmark, robustly across embedding spaces. Thus, we demonstrate a performance--diversity tradeoff in multilingual LLMs.
\end{abstract}

\section{Introduction}
In recent years, Large Language Models (LLMs) have become central in the field of Natural Language Processing (NLP). As their adoption increases, closer examination of the nature of the generated text is warranted. Specifically, empirical evaluations have revealed an artificial hivemind-like effect in open-ended generation \citep{jiang2025artificialhivemindopenendedhomogeneity}, defined by high levels of inter- and intra-model homogeneity, raising concerns about the long-term homogenization of text on the internet. However, the causes of the observed mode collapse remain unclear.

Potential explanations for this phenomenon range from inductive biases inherent to the optimization algorithms used, to the diversity and scale of the training distribution. In this work, we investigate the role of language on output diversity.

\begin{figure}[ht]
    \centering
    \includegraphics[width=0.85\linewidth]{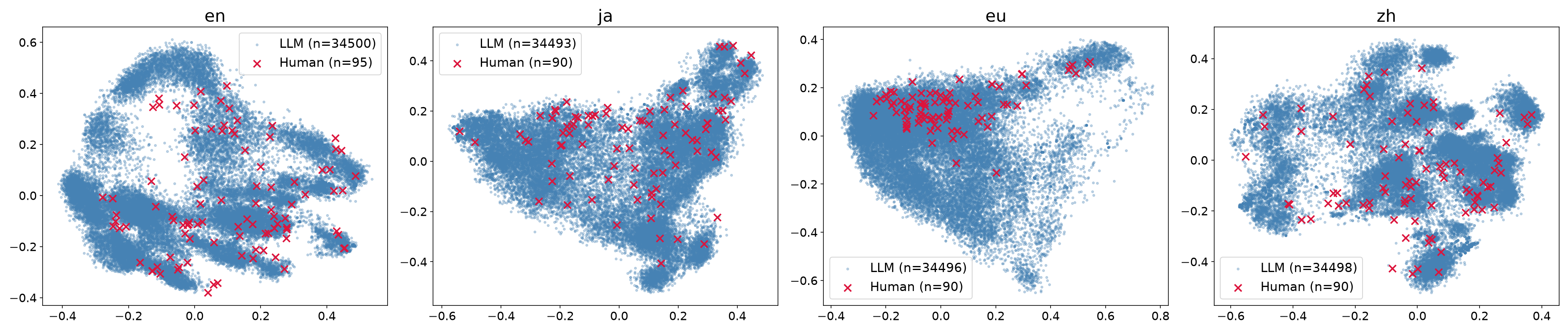}
    \caption{\textbf{Per-language Principal Component Analysis (PCA) projections of response embeddings.} For four representative languages (English, Japanese, Basque, and Chinese), blue point clouds represent the spatial distribution of pooled LLM generations, while red $\times$ markers display the human answers to the same prompts. LLM generations concentrate into dense modes, whereas the human answers are spread thinly over the same region. Each panel pools all 30 prompts of its language, so it shows the mode structure of the pooled distribution rather than the within-prompt gap measured in Section~\ref{sec:results}. Each of the two plotted components explains 5--10\% of the embedding variance in its panel.}
    \label{fig:image3}
\end{figure}

Our contributions can be summarized as follows:
\begin{itemize}
    \item We evaluate output diversity for 23 LLMs for 13 different languages.
    \item We show that the artificial hivemind effect is, in general, pervasive across languages.
    \item We show the intra-model effect is robust to how diversity is measured, replicating it in three embedding spaces and under an embedder-free lexical metric, and that both effects survive paired prompt-level tests with multiple-comparison correction.
    \item We show that the homogenization gap tracks how well a model handles a language, measured on a held-out translation benchmark, far more than properties of the language itself such as tokenization cost or morphology.
\end{itemize}

\section{Related Work}
Our work broadly relates to (i) language model evaluation, and (ii) the development of multilingual LLMs.

\subsection{Language Model Evaluation}
As language model architectures and capabilities have evolved, so have the evaluations we use to measure progress. Early evaluations of LLMs focused primarily on natural language processing and understanding tasks like question answering and text-summarization \citep{joshi-etal-2017-triviaqa, wang2019gluemultitaskbenchmarkanalysis, wang2020supergluestickierbenchmarkgeneralpurpose, rajpurkar-etal-2016-squad}, often relying on metrics like accuracy for single-answer tasks, and the BLEU or F1 score \citep{papineni-etal-2002-bleu} for open-ended tasks. As LLMs became more capable of reasoning-based tasks \cite{hendrycks2021measuringmassivemultitasklanguage} like solving math problems \citep{cobbe2021trainingverifierssolvemath, hendrycks2021measuringmathematicalproblemsolving, tsoukalas2024putnambenchevaluatingneuraltheoremprovers, glazer2025frontiermathbenchmarkevaluatingadvanced, ai-mathematical-olympiad-prize, ai-mathematical-olympiad-progress-prize-2} or generating code \cite{chen2021evaluatinglargelanguagemodels, jimenez2024swebenchlanguagemodelsresolve}, evaluations also evolved to measure these abilities. A parallel line of work has continued to evaluate LLM abilities on natural language tasks, with a focus on their ability to be creative \cite{eldan2023tinystoriessmalllanguagemodels, wu2025writingbenchcomprehensivebenchmarkgenerative, zheng2023judgingllmasajudgemtbenchchatbot}. However, many evaluations have highlighted the same gap: LLM-outputted text lacks diversity, and on open-ended tasks, tends to converge to similar clusters, even across vastly different families of models \cite{jiang2025artificialhivemindopenendedhomogeneity}. These evaluations have focused on LLM-generated English outputs without studying the effect of language on output diversity.

\begin{figure*}[t]
    \centering
    \begin{subfigure}[t]{0.245\textwidth}
        \centering
        \includegraphics[width=\linewidth,valign=t]{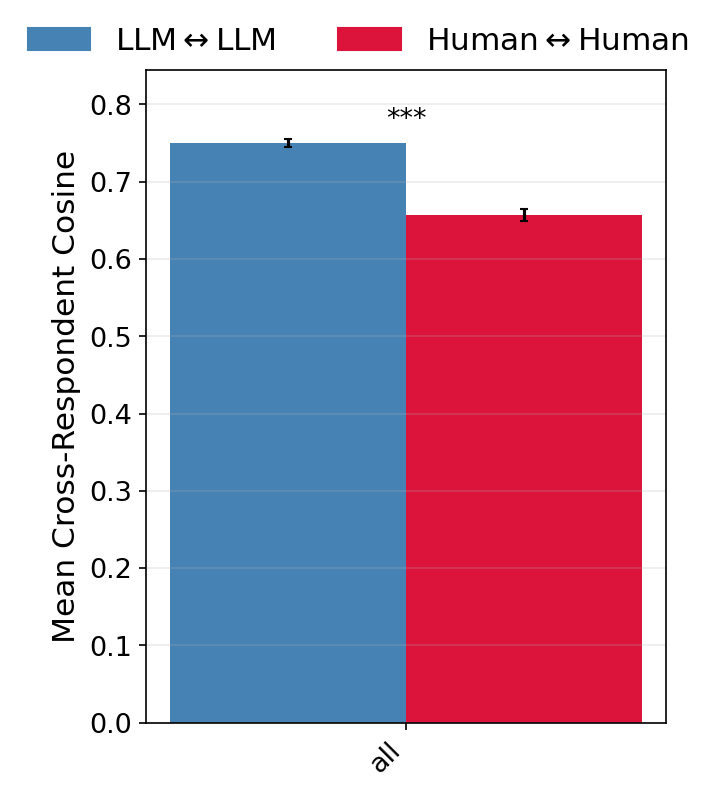}
        \vspace{15pt}
        \caption{General}
        \label{fig:f_tests_cross_bars_general}
    \end{subfigure}
    \hfill
    \begin{subfigure}[t]{0.690\textwidth}
        \centering
        \includegraphics[width=\linewidth,valign=t]{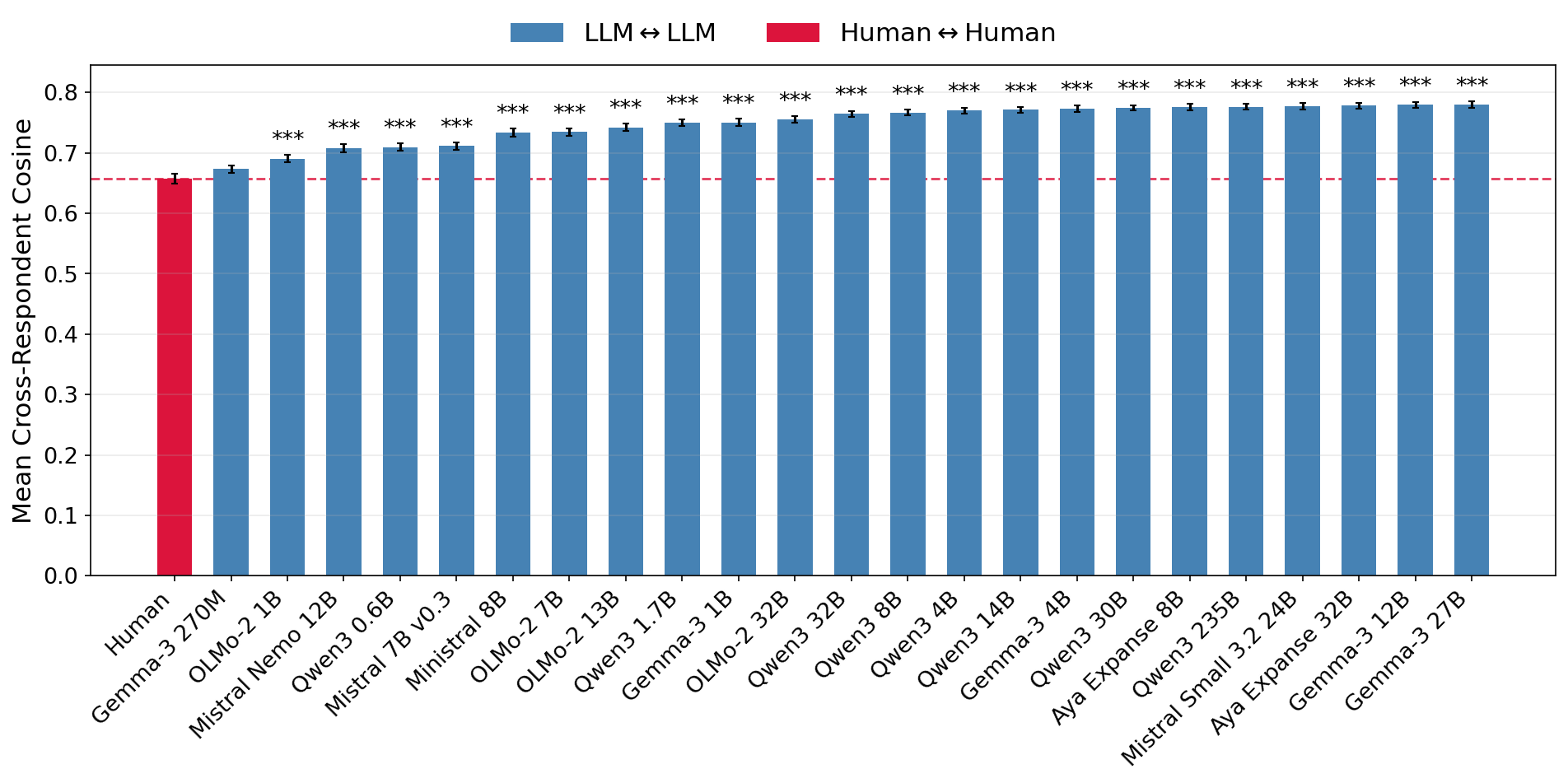}
        \vspace{-20pt}
        \caption{Per model}
        \label{fig:f_tests_cross_bars_model}
    \end{subfigure}

    \vspace{0.5em}
    \begin{subfigure}[t]{0.440\textwidth}
        \centering
        \includegraphics[width=\linewidth,valign=t]{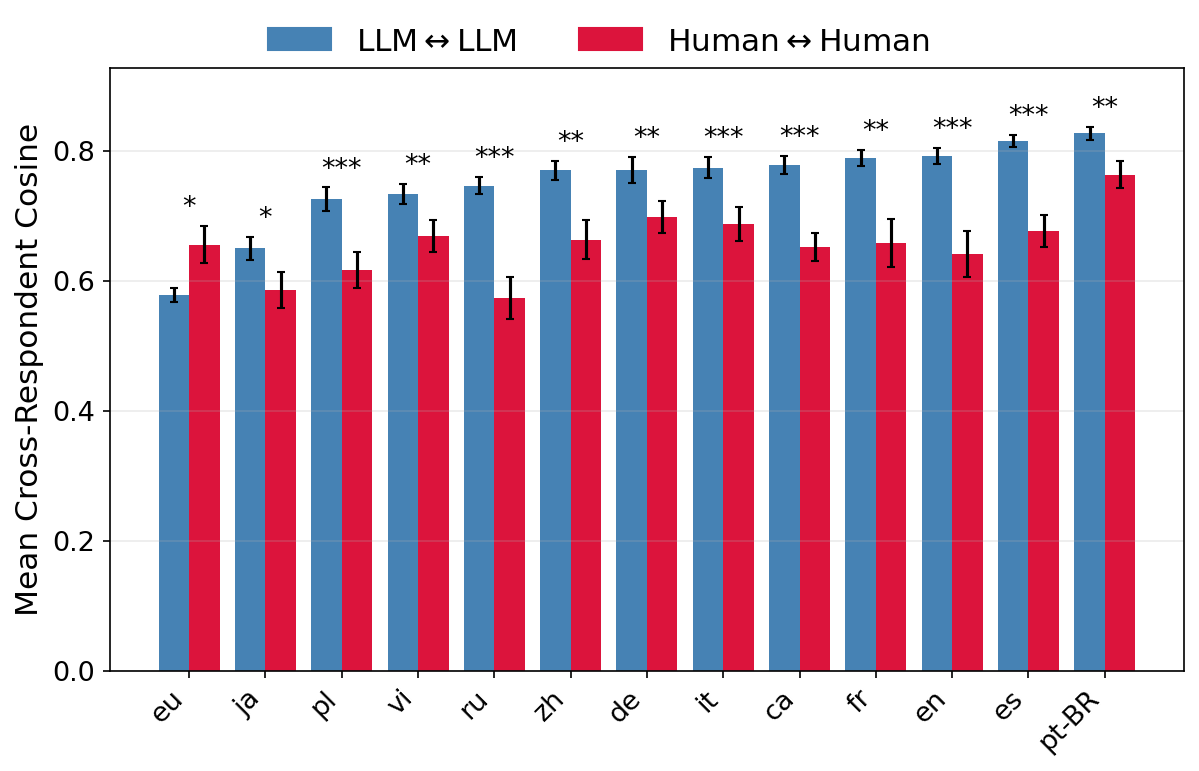}
        \vspace{25pt}
        \caption{Per language}
        \label{fig:f_tests_cross_bars_language}
    \end{subfigure}
    \hfill
    \begin{subfigure}[t]{0.500\textwidth}
        \centering
        \includegraphics[width=\linewidth,valign=t]{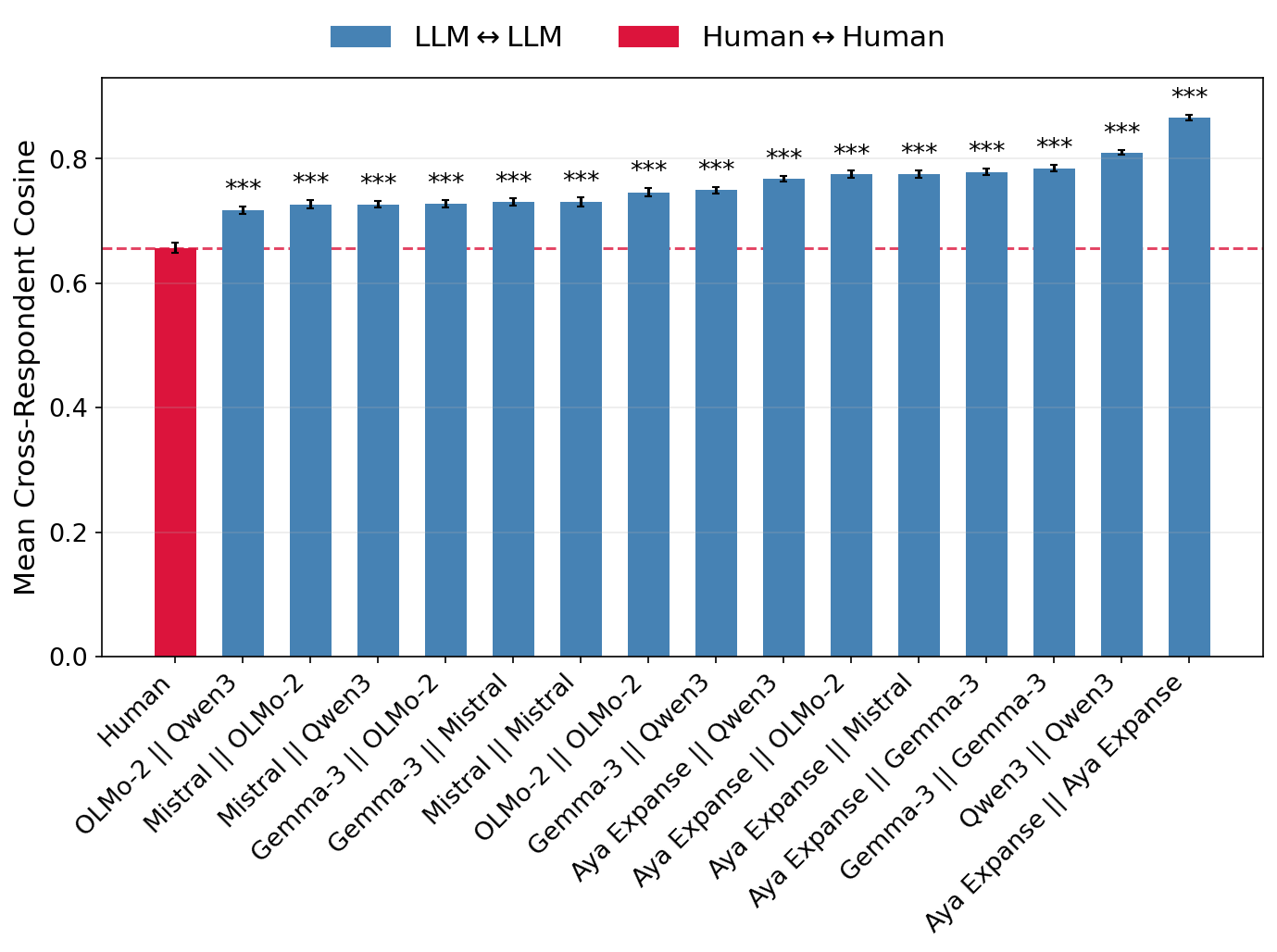}
        \vspace{-12pt}
        \caption{Per model-family pair}
        \label{fig:f_tests_cross_bars_family_pair}
    \end{subfigure}
    \caption{\textbf{Cross-respondent cosine similarities.} $LLM \leftrightarrow LLM$ vs. $Human \leftrightarrow Human$ evaluated (a) globally and (b) per individual model, and broken down (c) per language and (d) per model-family pair, on identical prompts. Higher LLM bars signify tighter cross-model clustering than humans exhibit with each other; in (b) and (d) the human baseline does not vary along the axis, so it appears once as the first bar, with a dashed reference line at its height. Whiskers are standard errors over prompts; stars mark Holm-corrected Wilcoxon signed-rank tests on the per-prompt differences (Appendix~\ref{app:stats}): $^{*}p<0.05$, $^{**}p<0.01$, $^{***}p<0.001$.}
    \label{fig:f_tests_cross_bars}
\end{figure*}

\subsection{Development of multilingual LLMs.}
Some of the first multilingual LLMs included Multilingual BERT \cite{pires-etal-2019-multilingual}, XLM-R \cite{conneau2020unsupervisedcrosslingualrepresentationlearning}, and BLOOM \cite{workshop2023bloom176bparameteropenaccessmultilingual}. More recent variants include Aya \cite{ustun-etal-2024-aya}, and national efforts to train LLMs primarily in a specific language \cite{sarvam2024sarvam1, ekgren2024gptsw3autoregressivelanguagemodel}. Even popular models like Qwen \cite{yang2025qwen3technicalreport} and Deepseek \cite{deepseekai2025deepseekv3technicalreport}, trained in China, are trained to perform tasks in both English and Mandarin. Despite these efforts, evaluations on parallel multilingual benchmarks still show that LLMs are generally less performant on non-English tasks \cite{xuan-etal-2025-mmlu, Bandarkar_2024}. This highlights the need for additional work in multilingual LLM development. 

In this work, we identify another gap in the evaluation of multilingual LLMs: their ability to output diverse text. For this, we use the OpenAssistant2 dataset, introduced as a multilingual large-scale alignment dataset for RLHF, to measure multilingual diversity in LLM outputs \cite{kopf2023openassistantconversationsdemocratizing}. 

\section{Methods}
\subsection{Dataset}
For our study, we use the OpenAssistant2 Conversations \cite{kopf2023openassistantconversationsdemocratizing}, a chat dataset spanning 135,174 messages in 28 different languages crowdsourced from 13,500 global volunteers. Because our analysis involves comparing diversity of human-generated responses to that of LLMs, we filter to prompts where 3 or more human responses are available, which narrows the dataset down to 13 languages where most prompts have 3 or more responses, over which we conduct our evaluation. The languages we evaluate are: English, Spanish, Russian, Chinese, German, French, Catalan, Brazilian Portuguese, Italian, Japanese, Basque, Polish, and Vietnamese.

\subsection{Models}
We evaluate models from five open-weight model families: OLMo 2, Qwen3, Gemma 3, Mistral and Aya Expanse. We choose models of a variety of sizes, ranging from 270M--235B, and architectures, including reasoning models and mixtures-of-experts. We use OpenAI’s \texttt{text-embedding-3-small} embedding model to measure embedding distances for our diversity metrics. 
The full list of models, with checkpoint revisions, precision, decoding settings and chat-template details, is provided in Appendix~\ref{app:models}. 

Because the geometry of an embedding space reflects the corpus it was trained on, a diversity metric read off a single embedder risks measuring that corpus rather than the generations. We therefore repeat the entire analysis in two further embedding spaces and under a metric that uses no embedding space at all; Section~\ref{sec:embedding_robustness} reports the result.

\subsection{Metrics}
We measure the diversity of generated text using cosine similarity. We define mean pairwise cosine similarity over embeddings $\{\mathbf{z}_i\}_{i=1}^n$ as
$$
\frac{1}{\binom{n}{2}}
\sum_{i<j}
\cos(\mathbf{z}_i,\mathbf{z}_j)
=
\frac{2}{n(n-1)}
\sum_{i<j}
\frac{\mathbf{z}_i^\top \mathbf{z}_j}{\|\mathbf{z}_i\|_2 \|\mathbf{z}_j\|_2}.
$$

Specifically, we define the following metrics:
\begin{enumerate}
    \item Intra-model similarity. For a given prompt and model, we compute the mean pairwise cosine similarity of sentence embeddings over multiple samples, averaged over all prompts in our dataset.
    
    \item Inter-model similarity. For a given prompt and pair of distinct models, the expected cosine similarity between one random sample from each model. With 50 samples per model we compute this expectation exactly over all $50 \times 50$ cross-model pairs and average over prompts.

    \item Lexical similarity (embedder-free). For a given prompt and model we compute the mean pairwise Jaccard similarity \citep{jaccard1912distribution} of character 4-gram sets, $\frac{2}{n(n-1)}\sum_{i<j} |G_i \cap G_j| / |G_i \cup G_j|$, where $G_i$ is the set of character 4-grams of sample $i$. We use character rather than word $n$-grams because \texttt{zh} and \texttt{ja} have no whitespace word boundaries, and a \emph{pairwise} rather than a pooled statistic (such as distinct-$n$ \citep{li2016diversity}) because a cell holds 50 LLM samples but only ${\sim}3$ human responses: pooled $n$-gram ratios shrink as the pool grows, so a pooled metric would report a large effect from sample count alone. This estimator has the same form as the cosine metrics above and involves no learned representation; it plays the role Self-BLEU \citep{zhu2018texygen} plays for English-only diversity benchmarks.
\end{enumerate}
\vspace{-0.5em}
\subsection{Experiments}
\label{sec:experiments}
\vspace{-0.5em}
\paragraph{Generation.} To evaluate the universality of the artificial hivemind effect across different linguistic distributions, we execute a large-scale multilingual generation sweep. Utilizing our filtered prompt set of 390 open-ended prompts across 13 languages from the OpenAssistant2 (OASST2) dataset, we generate output samples with the 23 selected open-weight models. 

For each individual prompt and model combination, we generate 50 independent response samples. To ensure consistency and align with prior work \cite{jiang2025artificialhivemindopenendedhomogeneity}, generations are produced using a temperature of 1.0, a \texttt{top\_p} threshold of 0.9, a maximum token limit of 2048, and a fixed random seed of 42. We do not vary the decoding configuration, because Jiang et al.~\cite{jiang2025artificialhivemindopenendedhomogeneity} report that even diversity-oriented decoding (min-$p$ sampling at temperature $2.0$) leaves 81\% of same-model response pairs above $0.7$ cosine similarity, so decoding can change the size of the effect but is unlikely to remove it. This setup results in a total of 448,500 target generations. Each generated text response is embedded using OpenAI’s text-embedding-3-small model, and the resulting vectors are $L^2$-normalized prior to metric computation.

\paragraph{Statistical tests.} To evaluate statistical significance across the general pool, model families, individual models, and languages, we compare LLM intra- and inter-model similarity against the human baseline with the prompt as the statistical unit. Each (language, prompt) cell contributes one LLM value (the mean pairwise cosine over a model's 50 samples, averaged over models where models are pooled) and one human value (over the ${\sim}3$ human answers), so the non-independent pairwise similarities within a prompt collapse into one observation, and the 50-versus-3 asymmetry affects each prompt's variance rather than its expectation. The F-tests we report are two-group one-way ANOVAs on these values (390 versus 390 for pooled and per-model tests, 30 versus 30 per language and per (model, language) cell). Because the two sides are paired by prompt and 299 cells are tested, Appendix~\ref{app:stats} repeats every comparison as a paired prompt-level analysis with Holm and Benjamini--Hochberg correction \citep{holm1979simple, benjamini1995controlling}. Every conclusion in this section holds under it, and all figure stars and whiskers are drawn from it.

\section{Results}
\label{sec:results}
Our experiments reveal that the artificial hivemind phenomenon is pervasive across languages, rejecting the hypothesis that mode collapse is solely a localized artifact of English-centric pre-training distributions or reinforcement learning alignment.

\subsection{Intra- and Inter-Model Clustering}
\label{sec:clustering}
\paragraph{Intra-model Clustering.} As demonstrated in Figure \ref{fig:f_tests_bars}, language models systematically generate text that clusters tighter than human responses across every evaluated language. We compare one LLM with many humans on purpose. Most people hear from many different humans but consult the same LLM, so a single model that is less diverse than a group of humans can homogenize the thinking of everyone who relies on it. The mean LLM intra-model cosine similarity reaches $0.857$, compared to the more diverse human baseline score of $0.657$ for the identical prompt set. The F-test confirms that the diversity gap is statistically significant with a value of $p < 10^{-88}$, and the paired prompt-level analysis of Appendix~\ref{app:stats} confirms it for all 23 models, all 13 languages and 271 of the 299 (model, language) cells after Holm correction (Table~\ref{tab:prompt_tests}). This sets one model, resampled, against ${\sim}3$ different human authors, since humans cannot be resampled (Section~\ref{sec:limitations}). The like-for-like comparison, different models against different humans, follows below and points the same way. The same-model similarity of $0.857$ also exceeds the different-model similarity of $0.750$, so resampling one model recovers less variety than switching models (Appendix~\ref{app:stats}).

To visually interpret the high-dimensional embedding spaces, we include Per-Language Principal Component Analysis (PCA) projections in Figure \ref{fig:image3}, where LLM generations collapse into high-density modes while the human answers are spread thinly over the same region of the embedding space. The within-prompt gap that the tests measure is not visible in a pooled projection. Appendix~\ref{app:pca} shows the same projection for all 13 languages.

\begin{figure}[t]
    \centering
    \includegraphics[width=0.83\textwidth]{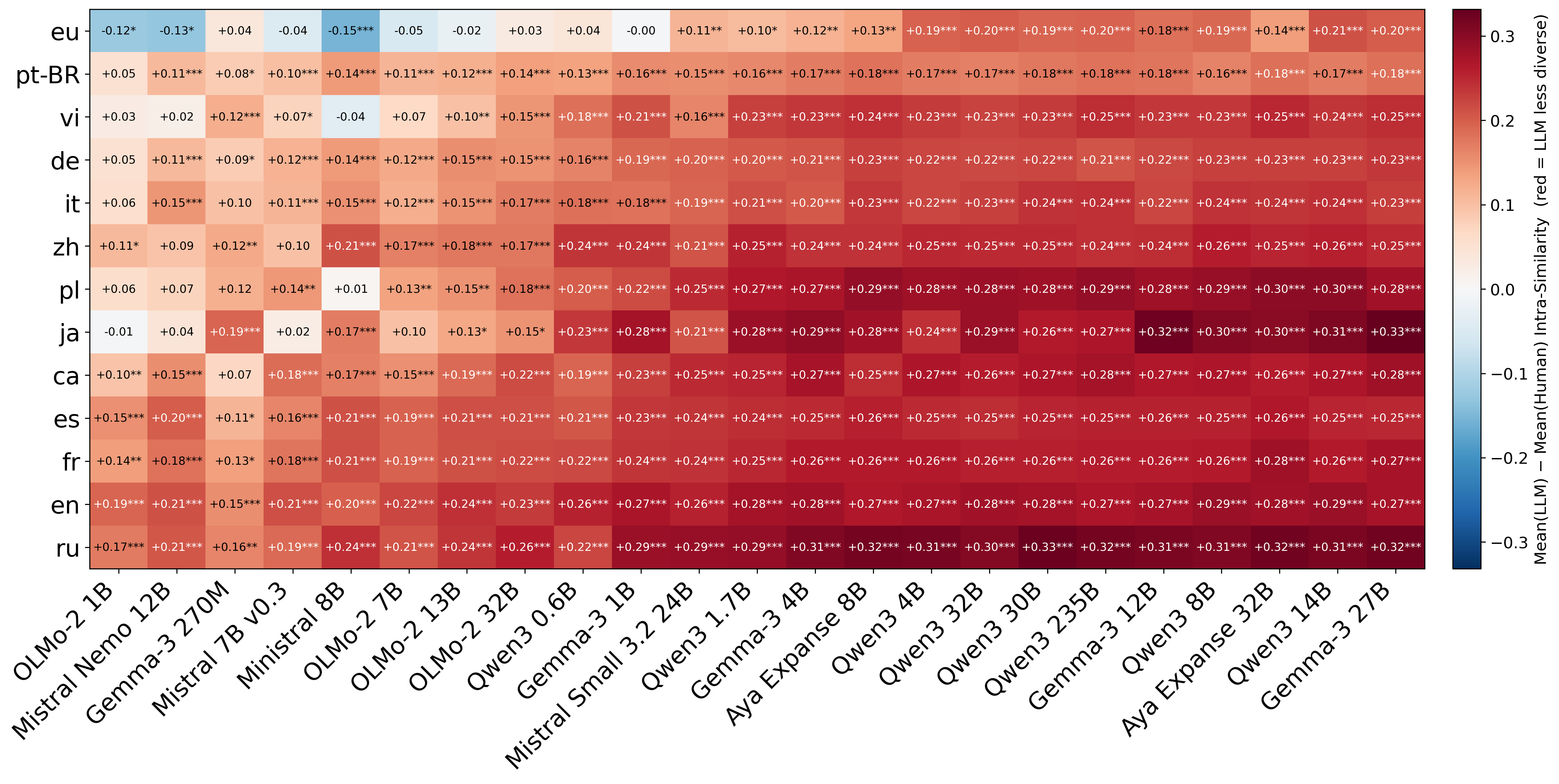}
    \vspace{-10pt}
        \caption{\textbf{Homogenization gap for each language $\times$ model pair.} Red tiles indicate high levels of homogenization compared to humans, while blue tiles indicate higher generative diversity. Stars as in Figure~\ref{fig:f_tests_cross_bars}, corrected across the 299 cells.}
    \label{fig:f_tests_heatmap_model_transposed}
\end{figure}

\paragraph{Inter-model Clustering.} To address whether shifting across distinct model choices injects meaningful structural variation into open-ended generations, we analyze cross-model behaviors via the inter-model similarity metric ($LLM \leftrightarrow LLM$). As shown in Figure \ref{fig:f_tests_cross_bars}, pairs of independent LLMs systematically agree more with each other than pairs of humans responding to the same prompt, across 12 out of the 13 evaluated languages (pooled gap $+0.093$, with every family pair and 12 of 13 languages significant after Holm correction, Table~\ref{tab:prompt_tests}). The exception to this trend occurs in Basque (\texttt{eu}), which is the only language in our data where human responses display tighter semantic clustering than that of the LLMs, and that reversal is itself significant. In contrast, for Japanese (\texttt{ja}) the model-to-human cosine similarity gap reaches its largest difference of $+0.33$, observed for Gemma-3 27B. The respondent--respondent similarity matrix (Figure \ref{fig:respondent_matrix_full}) shows the same pattern: most pairs of models are more similar to each other than the human--human baseline of $65.7$.

\begin{wrapfigure}{r}
{0.4\linewidth}
\vspace{-25pt}
\centering
    \includegraphics[width=\linewidth]{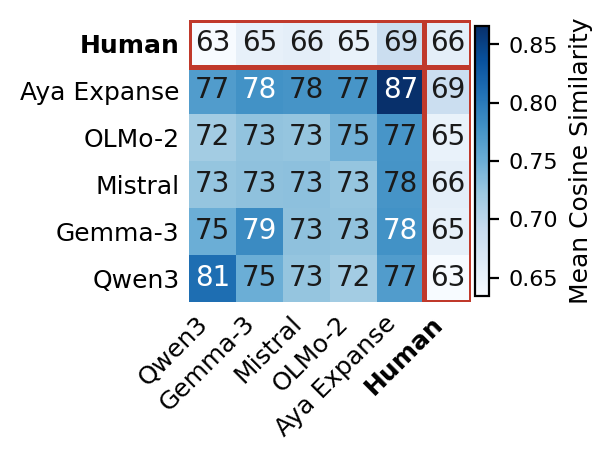}
    \caption{\textbf{Respondent--respondent similarity, aggregated by model family.} A \emph{respondent} is any source of an answer to a prompt: one of the 23 LLMs, or the pooled human annotators. Each cell is the mean pairwise cosine similarity between answers from two \emph{different} respondents to the same prompt, averaged over all prompts and languages ($\times 100$). 
    }
    \label{fig:respondent_matrix_family}
    \vspace{-10pt}
\end{wrapfigure}

\subsection{Model- and Family-specific Analysis}
To investigate the hivemind effect in more detail, we evaluate model specific diversity (Figure \ref{fig:f_tests_heatmap_model_transposed}), cross-family interaction (Figure \ref{fig:f_tests_cross_heatmap}, Appendix~\ref{app:robustness}) and respondent-to-respondent similarity (Figure \ref{fig:respondent_matrix_family}). A \emph{respondent} is one of the 23 LLMs or the pooled human annotators (24 in total); each cell is the mean similarity between two respondents' answers to the same prompt. The LLM diagonal resamples a single model at $T=1.0$, while the human diagonal compares the ${\sim}3$ different annotators per prompt, since humans cannot be resampled.

The model-by-language matrix (Figure \ref{fig:f_tests_heatmap_model_transposed}) exposes a possible tradeoff between downstream optimization and generative diversity. The OLMo-2 and Mistral families consistently demonstrate high text diversity. We speculate that these models are less heavily steered by hyper-alignment pipelines, resulting in the preservation of more diverse generations. In contrast, the Aya Expanse family \citep{dang2024ayaexpansecombiningresearch} is explicitly instruction-tuned for multilingual tasks, yet it provides less diverse responses across most languages. Similarly, despite heavy pre-training and alignment in Mandarin, Qwen models \cite{yang2025qwen3technicalreport} output responses with low diversity in Chinese (\texttt{zh}). This suggests the possibility that there is a trade-off between model performance and diversity.

Low-parameter footprints like Gemma-3 270M and Gemma-3 1B display high diversity in Figure \ref{fig:f_tests_heatmap_model_transposed} for languages like Japanese (\texttt{ja}) and Basque (\texttt{eu}). However, qualitative log inspection reveals this is a statistical artifact of generation failure. In non-English environments, the smaller models experience linguistic and syntactic breakdown, resulting in incoherent character strings artificially lowering the mean pairwise cosine similarity.

The inter-model homogenization is visible in Figures \ref{fig:respondent_matrix_family} and \ref{fig:f_tests_cross_heatmap}. The family-pair heatmap (Figure \ref{fig:f_tests_cross_heatmap}) displays higher than human homogeneity across almost all languages and families. This behavior is further illustrated by the respondent--respondent matrix (Figure \ref{fig:respondent_matrix_family}), which reveals higher mutual similarity between model families than that of human responses, especially for Qwen3 and Gemma-3. Of the $253$ distinct LLM--LLM model pairs, $245$ score above the human--human baseline of $65.7$, as do all $10$ family-level pairings (minimum $71.7$), indicating lower diversity. The eight exceptions all involve Gemma-3 270M or Qwen3 0.6B, the models whose non-English output degrades into incoherence, so they reflect generation failure rather than genuine diversity.

\subsection{Robustness to the Choice of Similarity Metric}
\label{sec:embedding_robustness}
All results above are computed in one embedding space, and the geometry of that space reflects the corpus it was trained on. An English-heavy training corpus could in principle place non-English generations closer together for reasons unrelated to their diversity. We therefore re-embed all 448{,}500 generations in two further spaces and recompute the intra-model metric there and under a similarity measure that uses no embedding space at all (Table~\ref{tab:robustness}). The inter-model comparison of Figure~\ref{fig:f_tests_cross_bars} is reported in the primary space only.

\begin{table}[t]
\centering
\caption{\textbf{The homogenization gap under three embedding spaces and an embedder-free lexical metric.} Mean similarity between two answers to the same prompt when they come from the same model resampled and from two different humans, averaged over all prompts and models, and the difference between them (the homogenization gap). \textbf{Langs.} counts languages with a positive gap; $\rho$ is the Spearman correlation of the per-cell gap against the primary metric. The effect is present in every language under every metric. The BGE-M3 500-char row re-embeds every response cut to its first 500 characters, on 10 of the 50 samples per prompt.}
\small
\setlength{\tabcolsep}{4pt}
\begin{tabular}{@{}lccccc@{}}
\toprule
\textbf{Similarity metric} & \textbf{Same model} & \textbf{Diff.\ humans} & \textbf{Gap} & \textbf{Langs.} & \textbf{$\rho$} \\
\midrule
text-embedding-3-small & 0.857 & 0.657 & +0.200 & 13/13 & --- \\
BGE-M3 & 0.870 & 0.727 & +0.143 & 13/13 & 0.92 \\
BGE-M3, 500-char cap & 0.857 & 0.723 & +0.134 & 13/13 & 0.91 \\
Qwen3-Embedding-0.6B & 0.866 & 0.666 & +0.200 & 13/13 & 0.93 \\
char 4-gram Jaccard & 0.275 & 0.133 & +0.141 & 13/13 & 0.71 \\
char 4-gram Jaccard, 500-char cap & 0.268 & 0.130 & +0.138 & 13/13 & 0.72 \\
\bottomrule
\end{tabular}
\label{tab:robustness}
\end{table}

The alternative embedders are chosen to differ in lineage. BGE-M3 \citep{chen2024m3} is an XLM-R encoder trained for multilingual retrieval, sharing no lineage with any evaluated model family; Qwen3-Embedding \citep{zhang2025qwen3} is decoder-derived and shares lineage with one. Both accept the longest generation without truncation.

Re-embedding with BGE-M3 reproduces the effect closely: the homogenization gap between LLMs and humans is positive in all 13 languages and in $285$ of the $299$ (model, language) cells, i.e.\ the $23 \times 13$ grid of Figure~\ref{fig:f_tests_heatmap_model_transposed}, and the per-cell gap correlates with the original at Spearman $\rho = 0.92$ ($98.3\%$ sign agreement). Its absolute magnitude is smaller ($+0.143$ vs.\ $+0.200$) because BGE-M3 places the human responses closer together ($0.727$ vs.\ $0.657$); the two spaces compress differently, but rank the same cells the same way.

The embedder-free metric is the most informative check, because agreement among embedders could still reflect a property all learned representations share. Under character 4-gram Jaccard the gap is $+0.141$, positive in all 13 languages, and agrees with the embedding metric on the sign of $97.7\%$ of the 299 (model, language) cells (Spearman $\rho = 0.71$ on the per-cell gap). Because LLM responses are roughly four times longer than human ones ($2337$ vs.\ $585$ characters on average) and lexical overlap is length-sensitive, we repeat it with every response capped at 500 characters; the gap barely moves ($+0.138$, $\rho = 0.96$ against the uncapped version), so verbosity is not driving the effect. Re-embedding the capped responses with BGE-M3 (on 10 of the 50 samples per prompt) gives the same answer in embedding space. The intra-model gap is $+0.134$ against $+0.143$ at full length, positive in all 13 languages (Table~\ref{tab:robustness}).

\section{What Predicts the Per-Language Gap?}
\label{sec:linguistic}
The hivemind effect holds everywhere, but not equally: in the primary embedding space the mean gap ranges from $0.076$ in Basque to $0.270$ in Russian. Only the low end of that range is stable. Basque has the smallest gap under all five metrics, while the ordering of the other twelve languages agrees well between embedders (Spearman $\rho = 0.76$--$0.85$) and not at all with the embedder-free ranking ($\rho = 0.38$, $p = 0.20$). We therefore ask which properties of a language, or of a model's command of it, account for the spread.

\paragraph{Predictors.} We test the four properties suggested by the structure of the problem, two varying per (model, language) and two per language. \textbf{Tokenization fertility} is the number of subword tokens a model's tokenizer spends on FLORES-200 devtest \citep{costa2022no}, relative to the same tokenizer's English count. Parallel text keeps the measure a property of the tokenizer rather than of the model's output distribution. \textbf{Model competence} is the chrF++ \citep{popovic2017chrf++} of the model's zero-shot translation of the 1,012 FLORES+ devtest sentences from English into the language, a held-out task that shares no prompts or outputs with the generations we score. It is undefined for English, which this section therefore omits. The GlotLID \citep{kargaran2023glotlid} target-language match rate over a cell's 1500 generations is used only as an outcome-quality filter (specification C) and as a fifth predictor in a comparison fit (specification F). \textbf{Resource level} is $\log_{10}$ of the language's Wikipedia article count. \textbf{Morphological complexity} is the type-token ratio of the FLORES parallel text; it is undefined for \texttt{zh} and \texttt{ja}, which lack whitespace word boundaries.

\paragraph{Method.} We regress the gap on all four predictors, $z$-scored, with \emph{crossed} random intercepts for model and for language. Crossing matters: resource level and morphology take only 13 distinct values, so a model-only random effect would treat 13 languages as hundreds of independent observations. Two models lack a benchmark score,\footnote{Qwen3 235B and Qwen3 30B could not be run on the benchmark hardware, so this section uses the remaining 21 models and, with English as the translation source, 12 languages. All other sections use all 23 models and 13 languages.} and one predictor is incomplete. Morphology is undefined for \texttt{zh} and \texttt{ja}, so the full specification (A) fits $21 \times 10 = 210$ cells; dropping morphology (B) recovers \texttt{zh} and \texttt{ja} and uses all $252$, and restricting B to cells with a GlotLID match rate of at least $0.80$ (C) leaves $230$. Specification F adds the match rate to B as a fifth predictor. No cell is excluded for any other reason.

\begin{figure}[t]
    \centering
    \begin{subfigure}[t]{0.33\linewidth}
        \centering
        \includegraphics[width=\linewidth]{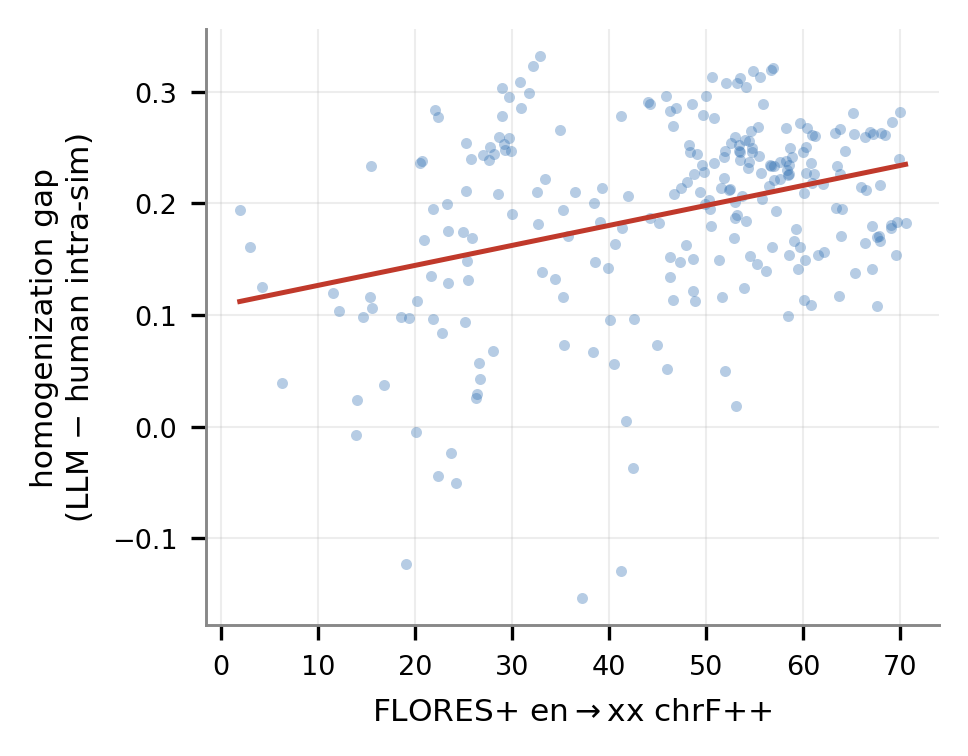}
        \caption{Gap vs.\ FLORES+ translation chrF++ under the primary embedder (Spearman $\rho = 0.27$, $p < 0.001$, $n = 252$).}
        \label{fig:linguistic_gap_competence}
    \end{subfigure}%
    \hfill%
    \begin{subfigure}[t]{0.425\linewidth}
        \centering
        \includegraphics[width=\linewidth]{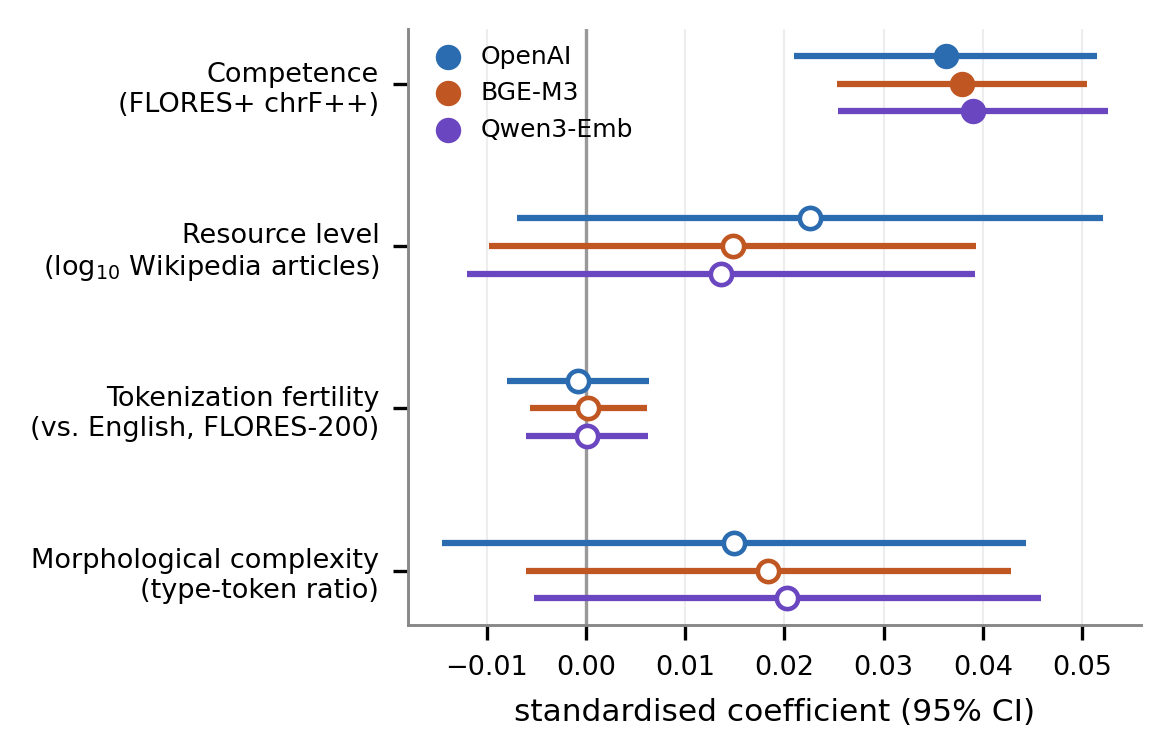}
        \caption{Standardised coefficients with 95\% CIs from the crossed-effects model, per embedding space (filled $= p<0.05$).}
        \label{fig:linguistic_gap_coefs}
    \end{subfigure}
    \caption{\textbf{Linguistic predictors of the homogenization gap.} (a) One point per (model, language) cell, the level the effect operates at; averaged per language it vanishes ($\rho = 0.06$, $p = 0.85$), since the effect lives across models within a language (specification H). (b) Fitted separately in each embedding space. Competence predicts the gap in all three spaces; resource level, fertility and morphology in none.}
    \label{fig:linguistic_gap}
\end{figure}

\paragraph{Results.} Fitting the model in all three embedding spaces separates one robust predictor from three that are not (Figure~\ref{fig:linguistic_gap}b, Table~\ref{tab:regression}). Competence predicts the gap in every space ($\beta = +0.036$/$+0.038$/$+0.039$ in specification A, $p < 0.001$ in every space for specifications A to C). Resource level ($p \geq 0.13$), tokenization fertility ($p \geq 0.10$) and morphological complexity ($p \geq 0.12$) do not predict it in any space. Resource level reaches $p < 0.05$ only when competence is standardised within each language (specification H, Appendix~\ref{app:linguistic}), which hands the between-language part of competence to the resource term, so we do not treat it as robust either way.

Homogenization thus tracks how well a model handles a language far more than the language's typology. Basque, the highest-fertility ($1.77\times$ English) and lowest-resource language in our set, and one of the three that models translate into worst, has the smallest gap under every metric we computed.

Four checks support this reading: dropping the 22 cells with a GlotLID match rate below $0.80$ (specification C) leaves the coefficient intact in all three spaces ($\beta = +0.044$/$+0.044$/$+0.043$), so degenerate generation is not what carries it; standardising competence within each language (specification H) leaves it intact as well, so differences in chrF++ between scripts do not carry it; resource level predicts LLM intra-similarity ($\rho = 0.52$) but not human ($\rho = 0.02$), placing the effect on the model side; and fertility predicts raw LLM intra-similarity but largely cancels in the gap, shifting the human side in the same direction. Adding the GlotLID match rate to B as a fifth predictor (F) leaves competence significant in every space ($\beta = +0.023$/$+0.029$/$+0.029$, $p \leq 0.003$) while the match rate is itself significant ($\beta = +0.024$/$+0.018$/$+0.018$, $p < 0.001$), so on-language output is associated with homogenization over and above competence.

This quantifies the performance--diversity tradeoff suggested in Section~\ref{sec:results} as an association: in our data, the better a model's command of a language, the \emph{less} diverse what it writes in it. Because the design is observational, we do not claim that improving competence causes the loss of diversity (Section~\ref{sec:limitations}).

\section{Limitations}
\label{sec:limitations}

\paragraph{Language sample.} Our original design called for the 10--20 largest languages by native-speaker count \citep{eberhard2026ethnologue}. To our knowledge, OpenAssistant2 is the only multilingual open-ended corpus with three or more human responses per prompt, and it fixes the 13 languages we study. Together they account for roughly $2.7$ billion native speakers, against roughly $3.9$ billion for the twenty largest, but four languages (Mandarin, Spanish, English, Portuguese) carry that figure: ten of the thirteen are European, and the set contains no language of South Asia, no Arabic, Turkish, Korean, or Indonesian. Basque is the only genuinely low-resource language, so the conclusion that the gap tracks competence rather than typology rests on a narrow typological base.

\paragraph{Non-parallel prompts.} OpenAssistant2 prompts are written natively in each language rather than translated, so per-language comparisons conflate the language with its prompt and topic distribution. Model-level results are unaffected (all models answer identical prompts), but part of the per-language variation may be topical.

\paragraph{Human baseline.} Each prompt carries ${\sim}3$ human responses by different authors, against 50 resamples of each LLM. Humans cannot be resampled, so the human intra-similarity mixes author variance with response variance. This shifts the level of the baseline rather than the sign of the effect, and the like-for-like inter-model test of Section~\ref{sec:clustering} points the same way.

\paragraph{Competence measure and causality.} Competence is translation quality from English under a surface metric, so English itself is not covered, chrF++ levels are not comparable across scripts (hence the language random effect and specification H), two models lack scores, and the Qwen3 models were run without reasoning. The association in Section~\ref{sec:linguistic} is not a causal estimate, since we manipulated neither competence, training data, alignment nor language coverage.

\paragraph{Length control.} The 500-character cap was applied to the lexical metric and to BGE-M3 on 10 of the 50 samples per prompt, not to the primary embedder. The intra-model gap is unchanged under the cap, but the cap is a crude form of matching, since it truncates 80\% of LLM responses but only 38\% of human ones and so sets largely complete human answers against LLM openings. Generations of human-like length would need a new sweep with a length instruction, since in 63\% of (model, prompt) cells none of the 50 samples is that short.

\paragraph{Metric stability.} The per-language ordering of the gap agrees between embedders but not with the embedder-free metric, and only Basque's position replicates under all five. Under BGE-M3 the Basque intra-model test loses significance ($p = 0.088$). The inter-model comparison is reported in the primary embedding space only.

\paragraph{Scope.} All generations use one decoding configuration (temperature $1.0$, top-$p$ $0.9$, at most 2048 tokens) and the five Qwen3 dense models ran in their default thinking mode (Appendix~\ref{app:models}), and the evaluated models are open-weight only. Our conclusions apply to these models, languages, prompts and decoding setting, and proprietary models and other sampling regimes remain untested.

\section{Conclusion}
For the 23 open-weight models, 13 languages and decoding configuration we study, the artificial hivemind phenomenon is in general pervasive and statistically significant across languages, model families and model sizes. The intra-model effect appears in all 13 languages under three different embedding models and under a lexical metric that uses none, so it is not an artifact of the embedding space, and both effects survive paired prompt-level tests corrected for multiple comparisons. Its magnitude is associated with how well a model handles a language, measured on a held-out benchmark, far more than with the language's typology, so the effect may grow rather than shrink as multilingual coverage improves. We encourage future work to model this performance--diversity tradeoff for multilingual tasks further, and to treat output diversity as an explicit objective that better language coverage alone does not provide.

\begin{ack}
All authors are supported by the Wallenberg AI, Autonomous
Systems and Software Program (WASP) funded by the Knut and Alice Wallenberg Foundation. The computations were enabled by resources provided by the National Academic Infrastructure for Supercomputing in Sweden (NAISS), partially funded by the Swedish Research Council through grant agreement no.\ 2022-06725.
\end{ack}

\bibliographystyle{plain}
\bibliography{anthology}

\newpage

\appendix

\begin{figure}[t]
    \centering
    \begin{subfigure}[b]{0.242\textwidth}
        \centering
        \includegraphics[width=\linewidth]{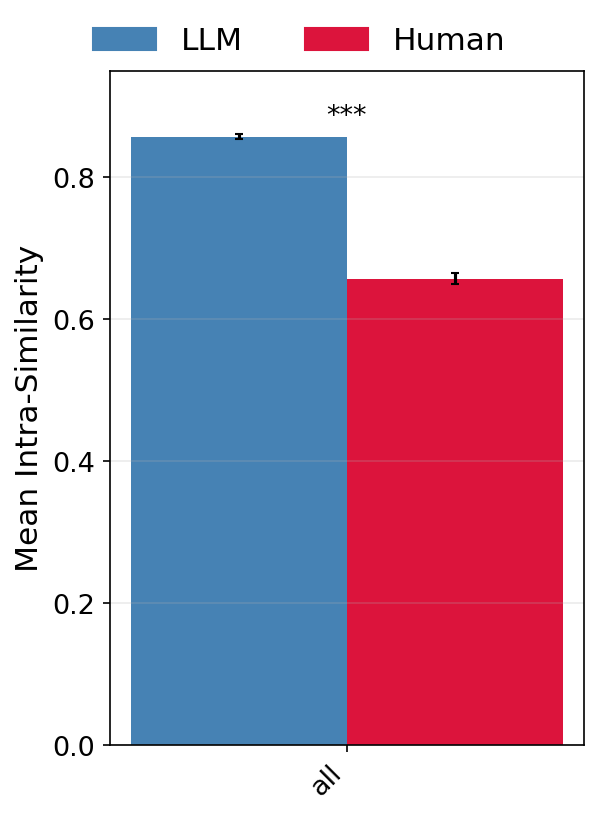}
        \caption{General}
        \label{fig:f_tests_bars_general}
    \end{subfigure}%
    \hfill%
    \begin{subfigure}[b]{0.232\textwidth}
        \centering
        \includegraphics[width=\linewidth]{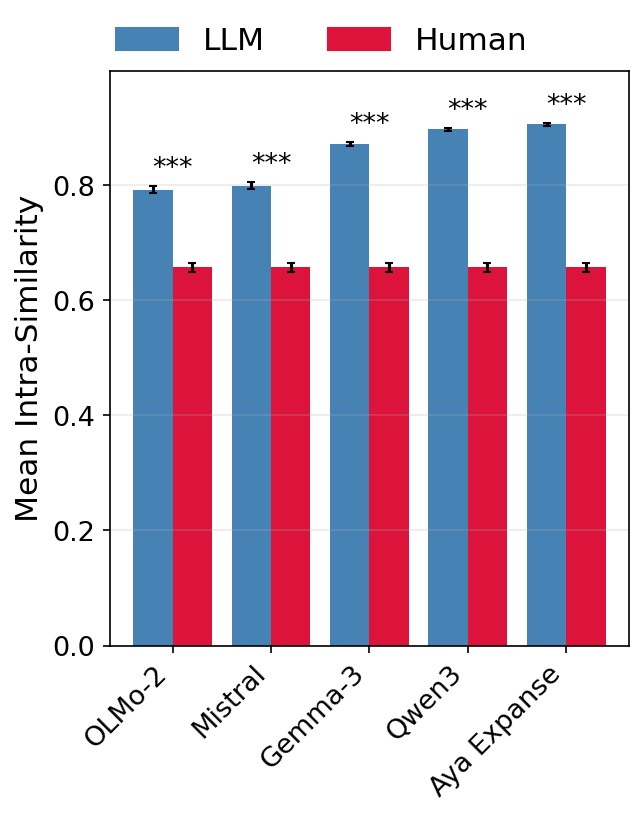}
        \caption{Per family}
        \label{fig:f_tests_bars_family}
    \end{subfigure}%
    \hfill%
    \begin{subfigure}[b]{0.466\textwidth}
        \centering
        \includegraphics[width=\linewidth]{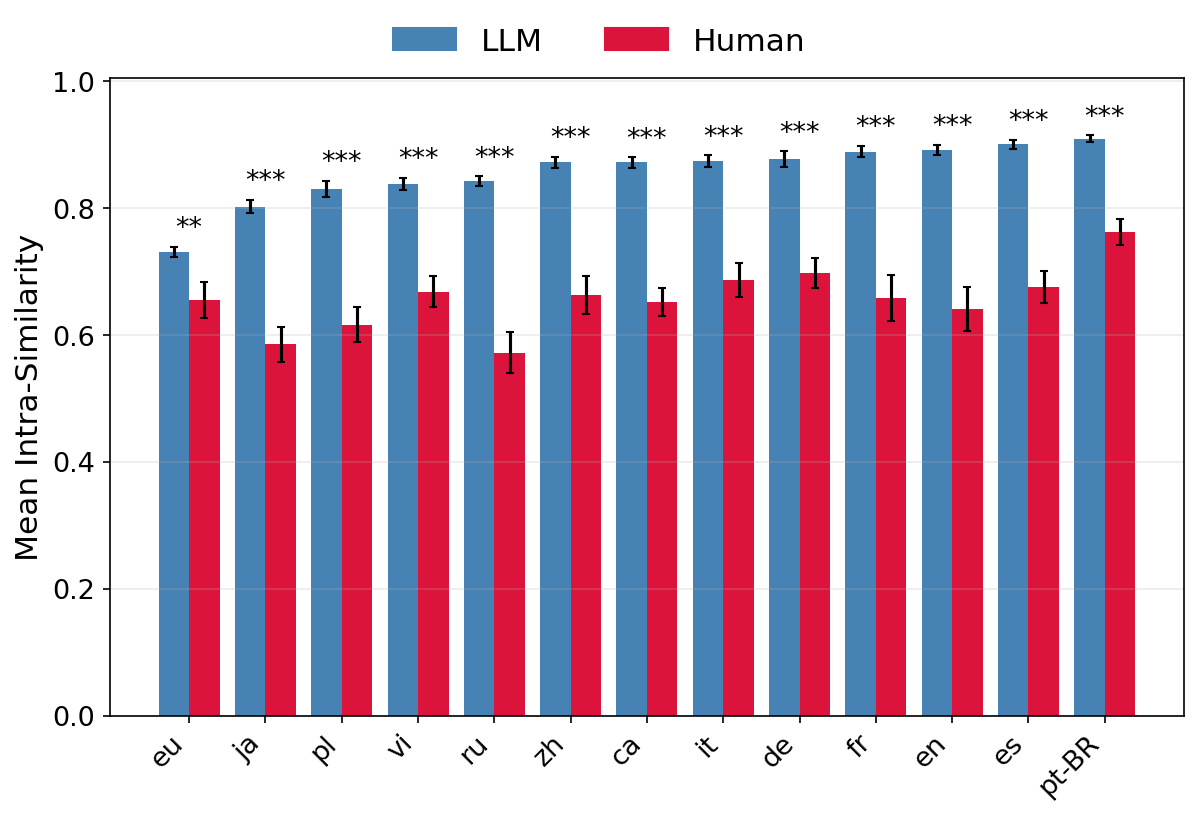}
        \caption{Per language}
        \label{fig:f_tests_bars_language}
    \end{subfigure}

    \vspace{0.7em}
    \begin{subfigure}[b]{\textwidth}
        \centering
        \includegraphics[width=0.95\linewidth]{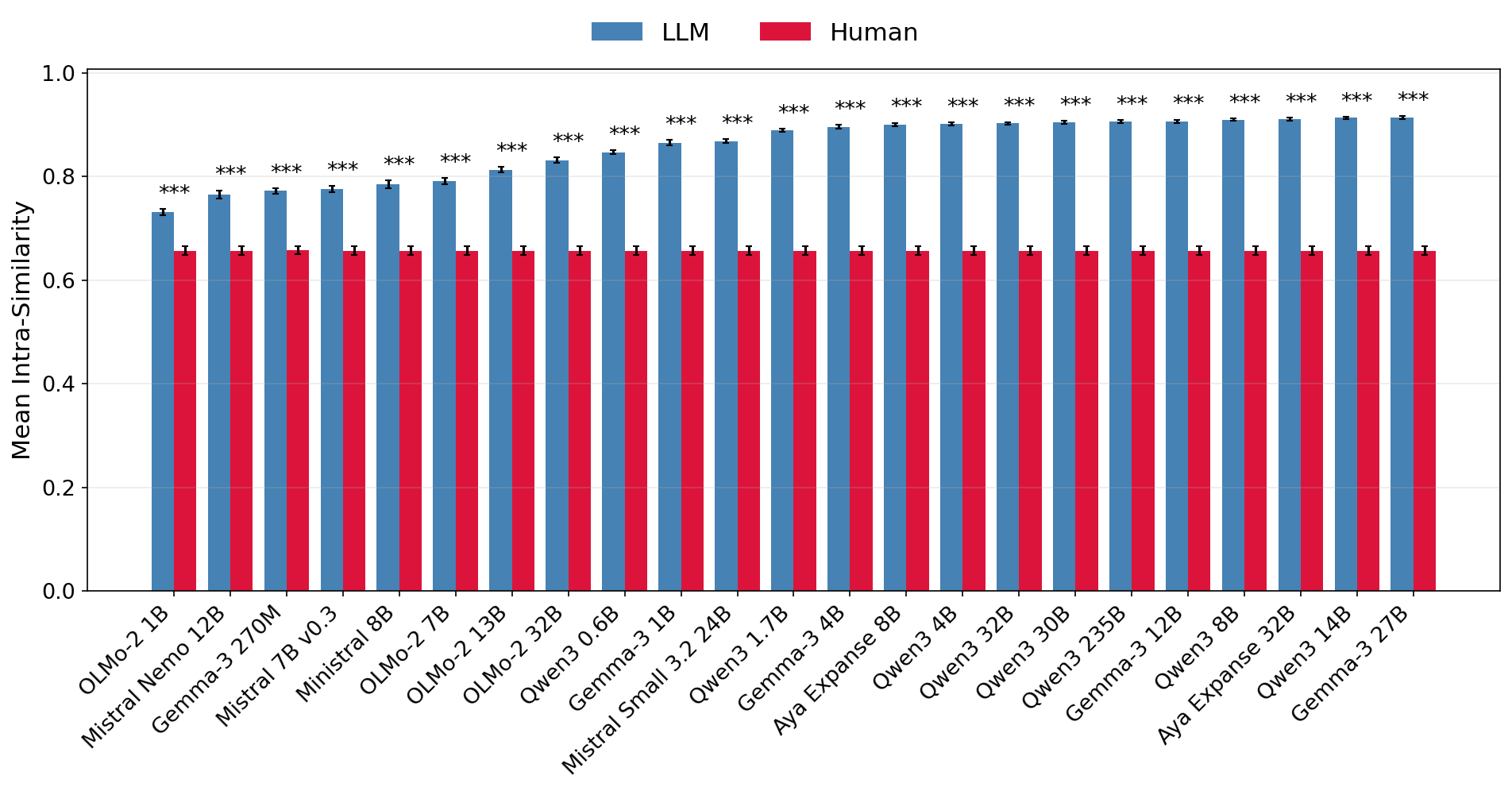}
        \caption{Per model}
        \label{fig:f_tests_bars_model}
    \end{subfigure}
    \caption{\textbf{LLM versus Human intra-similarity.} (a) Pooled globally, (b) per model family, (c) per language and (d) per individual model. Lower values signify higher generative diversity. Whiskers are standard errors over prompts, and stars mark Holm-corrected Wilcoxon signed-rank tests on the per-prompt differences (Appendix~\ref{app:stats}): $^{*}p<0.05$, $^{**}p<0.01$, $^{***}p<0.001$.}
    \label{fig:f_tests_bars}
\end{figure}

\section{Models and Compute}
\label{app:models}
In Table \ref{tab:model-families}, we include a list of all the model families and models we experiment with, together with the Hugging Face checkpoint and revision of each, the precision it was served at and the GPUs it ran on. Every checkpoint is the instruction-tuned release and was served at the precision declared in its config. Qwen3 235B is an FP8-quantized release.

\paragraph{Prompts and decoding.} The 30 prompts per language are OpenAssistant2 root prompts with at least three reviewed human replies, sampled with a fixed seed. The replies form the human baseline. All generations are produced with vLLM \citep{kwon2023efficient} through its chat interface, which applies each checkpoint's own chat template, with a single user turn and no system prompt. Generation stops at the template's end-of-turn token or at the 2048-token limit. The five Qwen3 dense checkpoints (0.6B--32B) default to thinking mode and emit a \texttt{<think>} block, which we keep as part of the response when embedding. For the GlotLID match rate used in Section~\ref{sec:linguistic} we strip the block first, and a block left unclosed because the token limit was hit counts as a non-match. Of the 448{,}500 generations, 79 are empty strings and are dropped before embedding.

\paragraph{Embedding and language identification.} \texttt{text-embedding-3-small} was queried through the OpenAI API in May 2026. The API exposes no snapshot identifier for embedding models. BGE-M3 \citep{chen2024m3} (revision \texttt{5617a9f6}) and Qwen3-Embedding-0.6B \citep{zhang2025qwen3} (revision \texttt{97b0c614}) were run with sentence-transformers at their native sequence limits, which are longer than any generation. Language identification uses GlotLID v3 \citep{kargaran2023glotlid} (revision \texttt{85cd6716}). For \texttt{de}, \texttt{ja} and \texttt{zh} we also count the closely related labels Swiss German, Hiragana-only Japanese and Wu/Cantonese as matches.

\paragraph{Compute.} All experiments are inference-only. Generations were produced with vLLM on a SLURM cluster using NVIDIA A40 (48\,GB) and A100 (80\,GB) GPUs; the full sweep fit within 130 GPU-hours. Embedding used the OpenAI API plus a few hours on a single NVIDIA RTX PRO 6000 Blackwell (96\,GB), and all downstream analysis runs on CPU. Code and all derived metrics are released at \url{https://github.com/xaverdavey/multilingual-artificial-hivemind}.

\begin{table*}[t]
\caption{\textbf{Breakdown of the 23 open-weight model variants across 5 families utilized in the multilingual homogeneity evaluation.} Revision is the Hugging Face commit that was the repository head when generation ran. Precision is the dtype declared in the checkpoint's config. Chars is the mean response length in characters, reasoning block included where present; Trunc.\ is the share of responses cut off by the 2048-token limit; Think marks checkpoints whose responses carry a \texttt{<think>} reasoning block.}
\centering
\scriptsize
\setlength{\tabcolsep}{3pt}
\resizebox{\textwidth}{!}{%
\begin{tabular}{llllrrrl}
\toprule
\textbf{Model} & \textbf{Checkpoint} & \textbf{Rev.} & \textbf{Prec.} & \textbf{GPUs} & \textbf{Chars} & \textbf{Trunc.\ \%} & \textbf{Think} \\
\midrule
Qwen3 0.6B & \texttt{Qwen/Qwen3-0.6B} & \texttt{c1899de2} & bf16 & 1$\times$ A40 48\,GB & 2,320 & 1.1 & yes \\
Qwen3 1.7B & \texttt{Qwen/Qwen3-1.7B} & \texttt{70d244cc} & bf16 & 1$\times$ A40 48\,GB & 4,442 & 14.8 & yes \\
Qwen3 4B & \texttt{Qwen/Qwen3-4B-Instruct-2507} & \texttt{cdbee75f} & bf16 & 1$\times$ A40 48\,GB & 2,546 & 7.2 & no \\
Qwen3 8B & \texttt{Qwen/Qwen3-8B} & \texttt{b968826d} & bf16 & 1$\times$ A40 48\,GB & 4,635 & 18.5 & yes \\
Qwen3 14B & \texttt{Qwen/Qwen3-14B} & \texttt{40c06982} & bf16 & 1$\times$ A40 48\,GB & 4,175 & 12.4 & yes \\
Qwen3 32B & \texttt{Qwen/Qwen3-32B} & \texttt{9216db57} & bf16 & 2$\times$ A100 80\,GB & 4,547 & 17.2 & yes \\
Qwen3 30B & \texttt{Qwen/Qwen3-30B-A3B-Instruct-2507} & \texttt{0d7cf239} & bf16 & 1$\times$ A100 80\,GB & 2,360 & 5.2 & no \\
Qwen3 235B & \texttt{Qwen/Qwen3-235B-A22B-Instruct-2507-FP8} & \texttt{e156cb4e} & FP8 & 4$\times$ A100 80\,GB & 2,188 & 3.4 & no \\
Gemma-3 270M & \texttt{google/gemma-3-270m-it} & \texttt{ac82b4e8} & bf16 & 1$\times$ A40 48\,GB & 1,522 & 3.9 & no \\
Gemma-3 1B & \texttt{google/gemma-3-1b-it} & \texttt{dcc83ea8} & bf16 & 1$\times$ A40 48\,GB & 2,332 & 0.8 & no \\
Gemma-3 4B & \texttt{google/gemma-3-4b-it} & \texttt{093f9f38} & bf16 & 1$\times$ A40 48\,GB & 2,460 & 0.5 & no \\
Gemma-3 12B & \texttt{google/gemma-3-12b-it} & \texttt{96b6f1ec} & bf16 & 1$\times$ A40 48\,GB & 2,584 & 0.5 & no \\
Gemma-3 27B & \texttt{google/gemma-3-27b-it} & \texttt{005ad340} & bf16 & 1$\times$ A100 80\,GB & 2,494 & 0.3 & no \\
Ministral 8B & \texttt{mistralai/Ministral-8B-Instruct-2410} & \texttt{2f494a19} & bf16 & 1$\times$ A40 48\,GB & 1,368 & 1.0 & no \\
Mistral 7B v0.3 & \texttt{mistralai/Mistral-7B-Instruct-v0.3} & \texttt{c170c708} & bf16 & 1$\times$ A40 48\,GB & 1,141 & 0.7 & no \\
Mistral Nemo 12B & \texttt{mistralai/Mistral-Nemo-Instruct-2407} & \texttt{04d8a905} & bf16 & 1$\times$ A40 48\,GB & 1,049 & 0.7 & no \\
Mistral Small 3.2 24B & \texttt{mistralai/Mistral-Small-3.2-24B-Instruct-2506} & \texttt{95a6d26c} & bf16 & 1$\times$ A100 80\,GB & 1,638 & 0.2 & no \\
OLMo-2 1B & \texttt{allenai/OLMo-2-0425-1B-Instruct} & \texttt{48d788ec} & bf16 & 1$\times$ A40 48\,GB & 1,543 & 0.0 & no \\
OLMo-2 7B & \texttt{allenai/OLMo-2-1124-7B-Instruct} & \texttt{470b1fba} & bf16 & 1$\times$ A40 48\,GB & 1,620 & 0.1 & no \\
OLMo-2 13B & \texttt{allenai/OLMo-2-1124-13B-Instruct} & \texttt{3a5c85ba} & bf16 & 1$\times$ A40 48\,GB & 1,629 & 0.0 & no \\
OLMo-2 32B & \texttt{allenai/OLMo-2-0325-32B-Instruct} & \texttt{b9602434} & bf16 & 2$\times$ A100 80\,GB & 1,563 & 0.0 & no \\
Aya Expanse 8B & \texttt{CohereLabs/aya-expanse-8b} & \texttt{5062468b} & fp16 & 1$\times$ A40 48\,GB & 1,851 & 0.0 & no \\
Aya Expanse 32B & \texttt{CohereLabs/aya-expanse-32b} & \texttt{b306ea27} & fp16 & 2$\times$ A100 80\,GB & 1,750 & 0.0 & no \\
\bottomrule
\end{tabular}}
\label{tab:model-families}
\end{table*}

\FloatBarrier
\section{Per-Language PCA Projections}
\label{app:pca}

\begin{figure}[t]
    \centering
    \includegraphics[width=\linewidth]{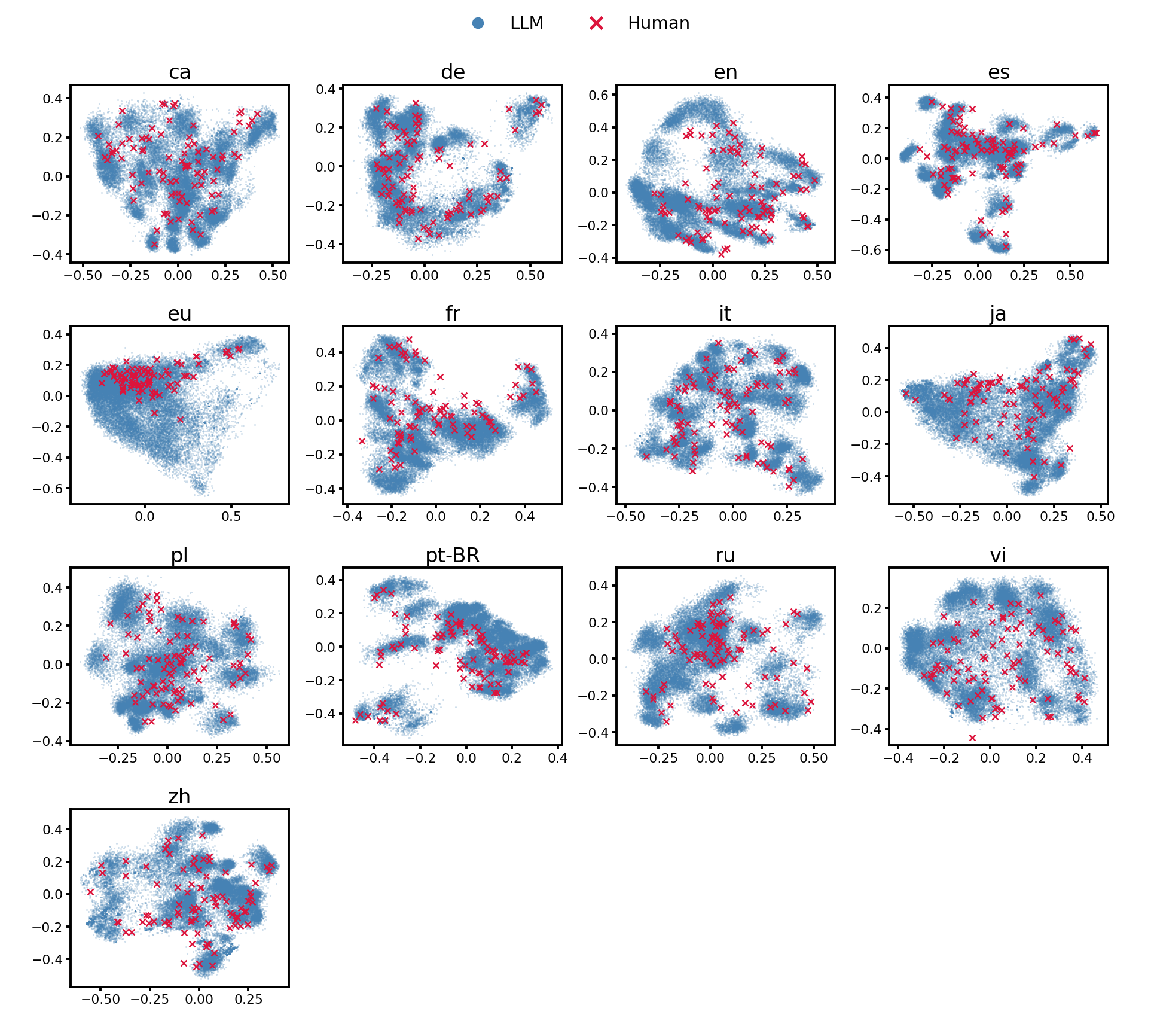}
    \caption{\textbf{Per-language PCA projections of response embeddings for all 13 languages.} Blue point clouds are pooled LLM generations; red $\times$ markers are human answers. Each of the two plotted components explains 5--10\% of the embedding variance in its panel.}
    \label{fig:pca_all}
\end{figure}

Figure~\ref{fig:pca_all} extends Figure~\ref{fig:image3} to every evaluated language. Each panel fits a fresh two-component PCA on that language's pooled points: the generations of all 23 LLMs (${\sim}34{,}500$ blue points per language) and the human answers (${\sim}90$ red $\times$ markers). The pattern of Figure~\ref{fig:image3} holds throughout: LLM generations collapse into high-density clusters while human answers spread across the occupied region.

\FloatBarrier
\section{Metric Robustness}
\label{app:robustness}

\begin{figure}[t]
    \centering
    \includegraphics[trim={0 0 0 20pt},clip,width=0.9\linewidth]{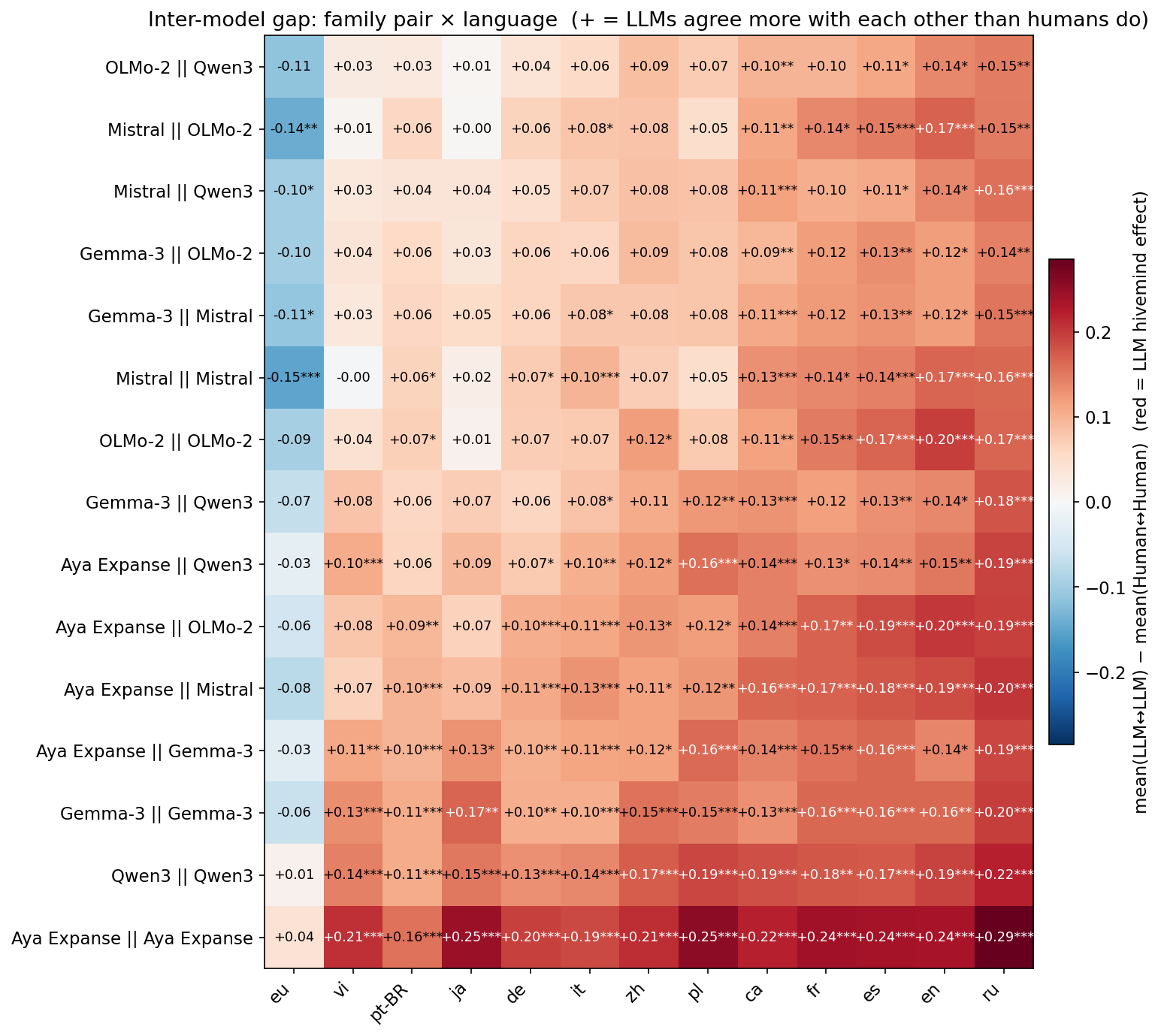}
    \caption{\textbf{Inter-model gap across family-pair combinations $\times$ languages.} Positive values (red blocks) indicate that the chosen model pairs cluster more tightly than human responses within that language. Stars mark Holm-corrected Wilcoxon signed-rank tests across the 195 cells: $^{*}p<0.05$, $^{**}p<0.01$, $^{***}p<0.001$.}
    \label{fig:f_tests_cross_heatmap}
\end{figure}

\begin{figure}[t]
    \centering
    \includegraphics[width=\linewidth]{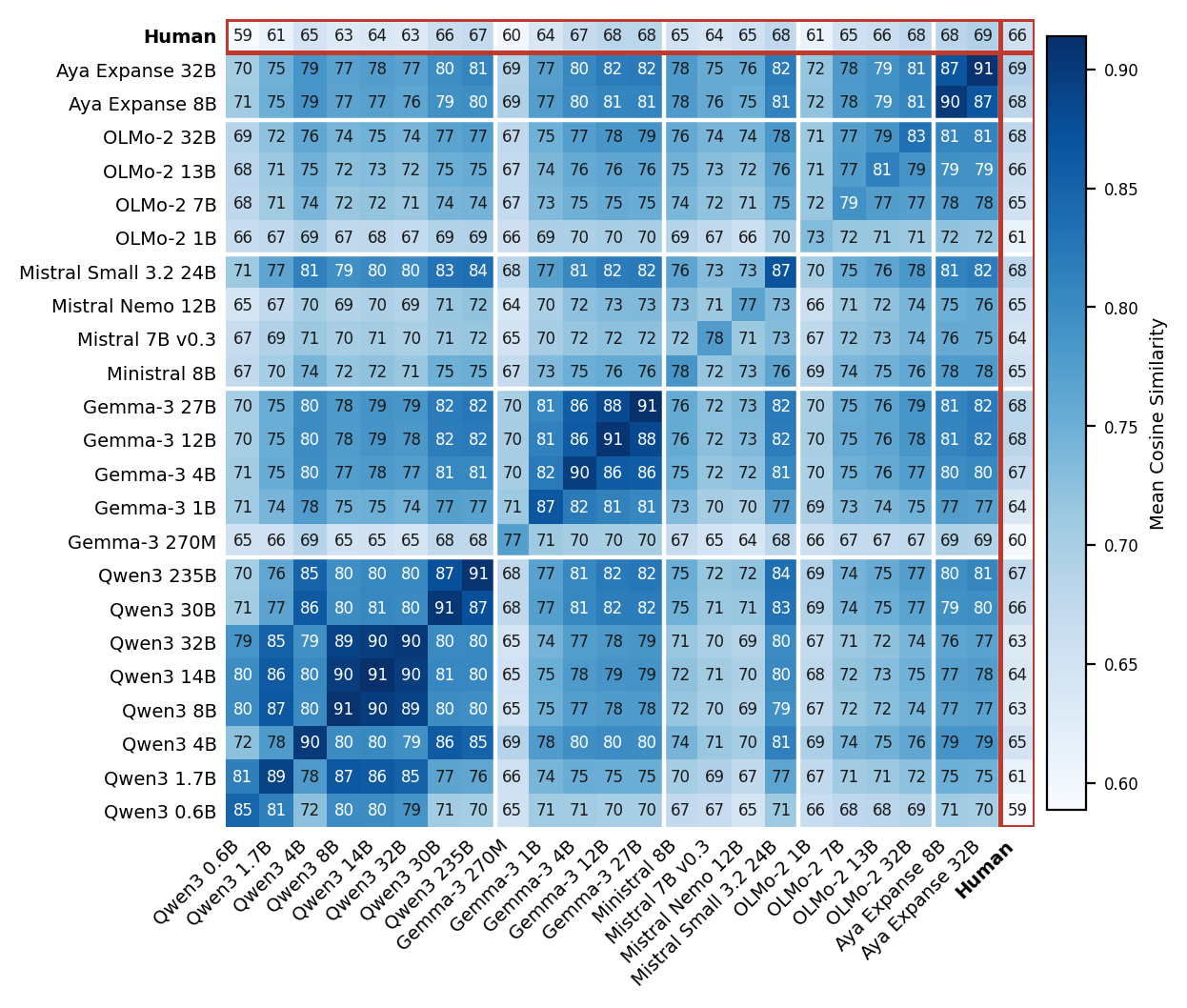}
    \caption{\textbf{Respondent--respondent similarity for all 24 respondents.} We report for each of the 23 LLMs, grouped by family, and the pooled human annotators (boxed), the mean pairwise cosine similarity between answers to the same prompt, averaged over all prompts and languages ($\times 100$). Off-diagonal cells compare two different respondents; the LLM diagonal resamples one model at $T=1.0$. Of the 253 distinct LLM--LLM pairs, 245 exceed the human--human value of 65.7; the eight that do not all involve Gemma-3 270M or Qwen3 0.6B, whose non-English output degenerates.}
    \label{fig:respondent_matrix_full}
\end{figure}

Table~\ref{tab:robustness_by_language} gives the per-language homogenization gap under every similarity metric. Figure~\ref{fig:f_tests_cross_heatmap} breaks the inter-model comparison of Figure~\ref{fig:f_tests_cross_bars} down by family pair and language, and Figure~\ref{fig:respondent_matrix_full} is the per-model respondent--respondent matrix summarised by family in Figure~\ref{fig:respondent_matrix_family}.

\begin{table*}[t]
\centering
\caption{\textbf{Per-language homogenization gap (LLM minus human intra-similarity) under each similarity metric, averaged over the 23 models.} Every language shows a positive gap under every metric.}
\small
\setlength{\tabcolsep}{3.5pt}
\begin{tabular}{lcccccc}
\toprule
\textbf{Lang.} & \textbf{BGE-M3} & \textbf{BGE-M3, 500-char cap} & \textbf{Qwen3-Emb} & \textbf{OpenAI} & \textbf{Jaccard} & \textbf{Jacc.\,500c} \\
\midrule
\texttt{en} & $+0.156$ & $+0.144$ & $+0.203$ & $+0.251$ & $+0.174$ & $+0.165$ \\
\texttt{de} & $+0.150$ & $+0.136$ & $+0.203$ & $+0.180$ & $+0.177$ & $+0.173$ \\
\texttt{fr} & $+0.165$ & $+0.155$ & $+0.212$ & $+0.231$ & $+0.149$ & $+0.148$ \\
\texttt{es} & $+0.155$ & $+0.145$ & $+0.230$ & $+0.225$ & $+0.182$ & $+0.172$ \\
\texttt{it} & $+0.158$ & $+0.146$ & $+0.191$ & $+0.187$ & $+0.166$ & $+0.161$ \\
\texttt{pt-BR} & $+0.099$ & $+0.084$ & $+0.168$ & $+0.147$ & $+0.140$ & $+0.122$ \\
\texttt{ca} & $+0.187$ & $+0.177$ & $+0.240$ & $+0.221$ & $+0.141$ & $+0.136$ \\
\texttt{pl} & $+0.170$ & $+0.164$ & $+0.214$ & $+0.213$ & $+0.139$ & $+0.143$ \\
\texttt{ru} & $+0.188$ & $+0.176$ & $+0.270$ & $+0.270$ & $+0.142$ & $+0.133$ \\
\texttt{vi} & $+0.103$ & $+0.090$ & $+0.159$ & $+0.170$ & $+0.147$ & $+0.144$ \\
\texttt{zh} & $+0.145$ & $+0.139$ & $+0.200$ & $+0.209$ & $+0.120$ & $+0.125$ \\
\texttt{ja} & $+0.140$ & $+0.135$ & $+0.209$ & $+0.217$ & $+0.160$ & $+0.173$ \\
\texttt{eu} & $+0.049$ & $+0.046$ & $+0.107$ & $+0.076$ & $+0.001$ & $+0.004$ \\
\bottomrule
\end{tabular}
\label{tab:robustness_by_language}
\end{table*}

\FloatBarrier
\section{Statistical Procedure and Prompt-Level Tests}
\label{app:stats}

The F-tests in the main text are two-group one-way ANOVAs on the per-prompt values defined in Section~\ref{sec:experiments}, with $(1, n_a + n_b - 2)$ degrees of freedom, where $n_a$ and $n_b$ are the numbers of prompts on the LLM and the human side. The two counts are equal except for Gemma-3 270M, which returned an empty string for all 50 samples of one German prompt, so that cell and that model's pooled test have one LLM value fewer. The statistic equals $t^2$ for the pooled-variance two-sample $t$-test on the same values. Because the LLM and human values are paired by prompt, all significance markers and error bars in the figures come from the paired analysis described below. Whiskers are standard errors over prompts, and stars are Holm-adjusted Wilcoxon signed-rank $p$-values ($^{*}p<0.05$, $^{**}p<0.01$, $^{***}p<0.001$).

For every comparison in the paper we form one difference per prompt, $\Delta_p = \text{LLM}_p - \text{Human}_p$, averaging the LLM side over the models or model pairs in the group first, and test $\Delta_p$ against zero with (i) a 95\% percentile bootstrap interval over prompts (10{,}000 resamples), (ii) the Wilcoxon signed-rank test \citep{wilcoxon1945individual}, and (iii) a sign-flip permutation test of the mean difference (20{,}000 Monte Carlo permutations). Within each level the $p$-values are adjusted across all groups tested at that level by Holm's step-down procedure \citep{holm1979simple}, which controls the family-wise error rate, and by Benjamini--Hochberg \citep{benjamini1995controlling}, which controls the false discovery rate. For the pooled gaps we also resample languages as clusters and then prompts within each selected language (10{,}000 resamples).

Table~\ref{tab:prompt_tests} summarises the result for the primary embedding space and, for the intra-model comparison, for the embedder-free metric with and without the 500-character cap and for BGE-M3 with the cap. The pooled intra-model gap is $+0.200$ (95\% interval $[0.186, 0.214]$ over prompts, $[0.169, 0.228]$ over language clusters) and the pooled inter-model gap is $+0.093$ ($[0.078, 0.108]$ and $[0.055, 0.126]$). The same-model versus different-model comparison of Section~\ref{sec:clustering}, one model resampled against two different models, gives $+0.107$ ($[0.102, 0.111]$) and is positive on all 390 prompts. The 28 (model, language) cells that fall short of a significant positive gap are concentrated in Basque, Japanese, Polish and Vietnamese and in the OLMo-2, Mistral 7B--12B and smallest Gemma-3 and Qwen3 checkpoints, matching Figure~\ref{fig:f_tests_heatmap_model_transposed}. Three of them, all in Basque, are significantly negative. Under each of the alternative metrics, Basque is the only language whose intra-model gap does not reach significance.

\begin{table*}[t]
\caption{\textbf{Paired prompt-level tests of the homogenization gap, with multiple-comparison correction.} Each group is tested on one LLM-minus-human difference per prompt ($n = 390$ prompts for per-model and per-family-pair groups, $30$ otherwise). Columns count the groups at that level ($k$) whose mean gap is positive, whose 95\% prompt-bootstrap interval lies above zero, and that are positive and significant at $0.05$ after Holm correction of the sign-flip permutation and Wilcoxon signed-rank $p$-values, and after Benjamini--Hochberg correction of the permutation $p$-values, each corrected across the $k$ groups of its row (a group that is significant in the reverse direction, such as Basque in the inter-model comparison, is not counted). The last row of the first block tests the same-model versus different-model comparison of Section~\ref{sec:clustering}, one model resampled against two different models, which needs no human baseline. The alternative metrics are reported for the intra-model comparison.}
\centering
\scriptsize
\setlength{\tabcolsep}{3pt}
\begin{tabular}{llrrrrrr}
\toprule
\textbf{Metric} & \textbf{Comparison / level} & $k$ & \textbf{gap}$>0$ & \textbf{CI}$>0$ & \textbf{perm.\ Holm} & \textbf{Wilc.\ Holm} & \textbf{perm.\ BH} \\
\midrule
text-embedding-3-small & Intra-model: per family & 5 & 5 & 5 & 5 & 5 & 5 \\
 & Intra-model: per model & 23 & 23 & 23 & 23 & 23 & 23 \\
 & Intra-model: per language & 13 & 13 & 13 & 13 & 13 & 13 \\
 & Intra-model: per (model, language) & 299 & 290 & 281 & 271 & 270 & 280 \\
 & Inter-model: per language & 13 & 12 & 12 & 12 & 12 & 12 \\
 & Inter-model: per model & 23 & 23 & 22 & 22 & 22 & 22 \\
 & Inter-model: per family pair & 15 & 15 & 15 & 15 & 15 & 15 \\
 & Inter-model: per (family pair, language) & 195 & 181 & 163 & 122 & 122 & 161 \\
 & Same vs.\ different models: per language & 13 & 13 & 13 & 13 & 13 & 13 \\
\midrule
char 4-gram Jaccard & Intra-model: per family & 5 & 5 & 5 & 5 & 5 & 5 \\
 & Intra-model: per model & 23 & 23 & 23 & 23 & 23 & 23 \\
 & Intra-model: per language & 13 & 13 & 12 & 12 & 12 & 12 \\
 & Intra-model: per (model, language) & 299 & 283 & 267 & 232 & 254 & 265 \\
\midrule
Jaccard, 500-char cap & Intra-model: per family & 5 & 5 & 5 & 5 & 5 & 5 \\
 & Intra-model: per model & 23 & 22 & 22 & 22 & 22 & 22 \\
 & Intra-model: per language & 13 & 13 & 12 & 12 & 12 & 12 \\
 & Intra-model: per (model, language) & 299 & 276 & 261 & 217 & 240 & 258 \\
\midrule
BGE-M3, 500-char cap & Intra-model: per family & 5 & 5 & 5 & 5 & 5 & 5 \\
 & Intra-model: per model & 23 & 22 & 22 & 22 & 22 & 22 \\
 & Intra-model: per language & 13 & 13 & 13 & 12 & 12 & 12 \\
 & Intra-model: per (model, language) & 299 & 281 & 272 & 244 & 243 & 272 \\
\bottomrule
\end{tabular}
\label{tab:prompt_tests}
\end{table*}

\FloatBarrier
\section{Linguistic Predictors}
\label{app:linguistic}

Table~\ref{tab:fertility} reports tokenization fertility per family, and Table~\ref{tab:regression} the four model specifications behind Section~\ref{sec:linguistic}.

\paragraph{Competence benchmark.} Each of the 21 models translated the 1,012 sentences of the FLORES+ devtest, the maintained successor of FLORES-200 \citep{costa2022no}, from English into each of the other 12 languages, zero-shot and at temperature 0, through the same vLLM chat interface as the main sweep, with an English instruction to reply with the translation only and with thinking disabled for the Qwen3 models. Competence is corpus-level chrF++ against the reference. It ranges from $1.9$ to $70.6$ across cells (median $50.3$) and is nearly unrelated to the GlotLID match rate (Spearman $\rho = 0.15$). Specification H refits B and F with competence standardised within each language, so that no difference between languages, such as the dependence of chrF++ on script, can enter. Competence stays significant ($\beta = +0.034$/$+0.032$/$+0.034$ in B and $+0.023$/$+0.025$/$+0.026$ in F, all $p < 0.001$), and resource level reaches $p < 0.05$ in H alone ($\beta = +0.031$/$+0.025$/$+0.024$ in B). As a placebo, regressing the human intra-similarity on the same predictors at the language level ($n = 12$) gives a positive competence coefficient ($\beta = +0.037$, $p = 0.012$ under the primary embedder; $p = 0.25$ and $0.06$ in the others), so languages that models translate into well have slightly more similar human answers, which would shrink the gap rather than inflate it.

\begin{table*}[t]
\centering
\caption{\textbf{Tokenization fertility on FLORES-200 devtest.} We report subword tokens each family's tokenizer spends per parallel sentence, relative to its own English count. Basque and Polish cost the most; Chinese the least.}
\small
\setlength{\tabcolsep}{4pt}
\begin{tabular}{lccccccccccccc}
\toprule
\textbf{Family} & \texttt{en} & \texttt{de} & \texttt{fr} & \texttt{es} & \texttt{it} & \texttt{pt-BR} & \texttt{ca} & \texttt{pl} & \texttt{ru} & \texttt{vi} & \texttt{zh} & \texttt{ja} & \texttt{eu} \\
\midrule
Aya Expanse & 1.00 & 1.32 & 1.35 & 1.26 & 1.32 & 1.16 & 1.58 & 1.34 & 1.27 & 1.33 & 1.04 & 1.22 & 1.76 \\
Gemma-3 & 1.00 & 1.33 & 1.41 & 1.30 & 1.38 & 1.26 & 1.54 & 1.51 & 1.37 & 1.37 & 1.08 & 1.21 & 1.68 \\
Mistral & 1.00 & 1.40 & 1.42 & 1.41 & 1.45 & 1.37 & 1.47 & 1.65 & 1.24 & 1.84 & 1.12 & 1.33 & 1.58 \\
OLMo-2 & 1.00 & 1.58 & 1.63 & 1.54 & 1.64 & 1.48 & 1.72 & 1.91 & 2.45 & 2.43 & 1.86 & 2.25 & 1.88 \\
Qwen3 & 1.00 & 1.56 & 1.58 & 1.52 & 1.63 & 1.45 & 1.69 & 1.76 & 1.75 & 1.44 & 1.00 & 1.40 & 1.87 \\
\bottomrule
\end{tabular}
\label{tab:fertility}
\end{table*}

\begin{table*}[t]
\centering
\caption{\textbf{Standardised coefficients from the crossed-effects model of the homogenization gap, fitted in each embedding space.} Competence is FLORES+ English$\to$target translation chrF++, a held-out benchmark that shares no prompts or outputs with the similarity scores; English is excluded, since the translations are from English. Spec A uses all four predictors on the languages for which morphology is defined (210 cells; the type-token ratio is undefined for \texttt{zh}/\texttt{ja}); B drops morphology to recover them (252 cells); C repeats B on the 230 cells whose generations have a GlotLID target-language match rate $\geq 0.80$, so degenerate output cannot drive the fit; F adds that match rate to B as a fifth predictor, testing whether it carries signal beyond competence. Competence is positive and significant ($p \leq 0.003$) in every specification and embedding space, including F, where the GlotLID match rate is also significant. Resource level, tokenization fertility and morphological complexity are significant in none. $^{*}p<0.05$, $^{***}p<0.001$.}
\footnotesize
\setlength{\tabcolsep}{3.5pt}
\begin{tabular}{llccccc}
\toprule
\textbf{Space} & \textbf{Spec} & \textbf{Competence} & \textbf{Resource level} & \textbf{Tokenization fertility} & \textbf{Morph. complexity} & \textbf{GlotLID} \\
\midrule
OpenAI & A & $+0.036^{***}$ & $+0.023$ & $-0.001$ & $+0.015$ & --- \\
 & B & $+0.038^{***}$ & $+0.020$ & $-0.006$ & --- & --- \\
 & C & $+0.044^{***}$ & $+0.011$ & $-0.004$ & --- & --- \\
 & F & $+0.023^{*}$ & $+0.019$ & $-0.002$ & --- & $+0.024^{***}$ \\
\midrule
BGE-M3 & A & $+0.038^{***}$ & $+0.015$ & $+0.000$ & $+0.018$ & --- \\
 & B & $+0.040^{***}$ & $+0.013$ & $-0.002$ & --- & --- \\
 & C & $+0.044^{***}$ & $+0.007$ & $-0.001$ & --- & --- \\
 & F & $+0.029^{***}$ & $+0.012$ & $+0.000$ & --- & $+0.018^{***}$ \\
\midrule
Qwen3-Emb & A & $+0.039^{***}$ & $+0.014$ & $+0.000$ & $+0.020$ & --- \\
 & B & $+0.041^{***}$ & $+0.012$ & $-0.003$ & --- & --- \\
 & C & $+0.043^{***}$ & $+0.007$ & $-0.002$ & --- & --- \\
 & F & $+0.029^{***}$ & $+0.011$ & $-0.001$ & --- & $+0.018^{***}$ \\
\bottomrule
\end{tabular}
\label{tab:regression}
\end{table*}

\FloatBarrier

\section*{NeurIPS Paper Checklist}

The checklist is designed to encourage best practices for responsible machine learning research, addressing issues of reproducibility, transparency, research ethics, and societal impact. Do not remove the checklist: {\bf The papers not including the checklist will be desk rejected.} The checklist should follow the references and follow the (optional) supplemental material.  The checklist does NOT count towards the page
limit. 

Please read the checklist guidelines carefully for information on how to answer these questions. For each question in the checklist:
\begin{itemize}
    \item You should answer \answerYes{}, \answerNo{}, or \answerNA{}.
    \item \answerNA{} means either that the question is Not Applicable for that particular paper or the relevant information is Not Available.
    \item Please provide a short (1--2 sentence) justification right after your answer (even for \answerNA). 
\end{itemize}

{\bf The checklist answers are an integral part of your paper submission.} They are visible to the reviewers, area chairs, senior area chairs, and ethics reviewers. You will also be asked to include it (after eventual revisions) with the final version of your paper, and its final version will be published with the paper.

The reviewers of your paper will be asked to use the checklist as one of the factors in their evaluation. While \answerYes{} is generally preferable to \answerNo{}, it is perfectly acceptable to answer \answerNo{} provided a proper justification is given (e.g., error bars are not reported because it would be too computationally expensive'' or ``we were unable to find the license for the dataset we used''). In general, answering \answerNo{} or \answerNA{} is not grounds for rejection. While the questions are phrased in a binary way, we acknowledge that the true answer is often more nuanced, so please just use your best judgment and write a justification to elaborate. All supporting evidence can appear either in the main paper or the supplemental material, provided in appendix. If you answer \answerYes{} to a question, in the justification please point to the section(s) where related material for the question can be found.

IMPORTANT, please:
\begin{itemize}
    \item {\bf Delete this instruction block, but keep the section heading ``NeurIPS Paper Checklist"},
    \item  {\bf Keep the checklist subsection headings, questions/answers and guidelines below.}
    \item {\bf Do not modify the questions and only use the provided macros for your answers}.
\end{itemize}


\begin{enumerate}

\item {\bf Claims}
    \item[] Question: Do the main claims made in the abstract and introduction accurately reflect the paper's contributions and scope?
    \item[] Answer: \answerYes{} 
    \item[] Justification: Abstract and methods/results go hand-in-hand.

\item {\bf Limitations}
    \item[] Question: Does the paper discuss the limitations of the work performed by the authors?
    \item[] Answer: \answerYes{} 
    \item[] Justification: Section 6 is a dedicated Limitations section.

\item {\bf Theory assumptions and proofs}
    \item[] Question: For each theoretical result, does the paper provide the full set of assumptions and a complete (and correct) proof?
    \item[] Answer: \answerNA{} 
    \item[] Justification: The paper contains no theoretical results.

    \item {\bf Experimental result reproducibility}
    \item[] Question: Does the paper fully disclose all the information needed to reproduce the main experimental results of the paper to the extent that it affects the main claims and/or conclusions of the paper (regardless of whether the code and data are provided or not)?
    \item[] Answer: \answerYes{} 
    \item[] Justification: Section 3 gives the dataset, models, metrics and decoding settings; Appendix A lists every checkpoint with its Hugging Face revision, precision, hardware, chat-template handling and truncation statistics, together with the embedding and language-identification model versions; Appendix D defines every statistical test.

\item {\bf Open access to data and code}
    \item[] Question: Does the paper provide open access to the data and code, with sufficient instructions to faithfully reproduce the main experimental results, as described in supplemental material?
    \item[] Answer: \answerYes{} 
    \item[] Justification: Code, the prompt-selection procedure and all derived metrics are released at the URL given in Appendix A.

\item {\bf Experimental setting/details}
    \item[] Question: Does the paper specify all the training and test details (e.g., data splits, hyperparameters, how they were chosen, type of optimizer) necessary to understand the results?
    \item[] Answer: \answerYes{} 
    \item[] Justification: All details are included in the methods section.

\item {\bf Experiment statistical significance}
    \item[] Question: Does the paper report error bars suitably and correctly defined or other appropriate information about the statistical significance of the experiments?
    \item[] Answer: \answerYes{} 
    \item[] Justification: The F-tests in the main text are defined in the Methods, with the prompt as the statistical unit. Every star and whisker in the figures comes from the paired prompt-level analysis of Appendix~\ref{app:stats}, where whiskers are standard errors over prompts and stars are Holm-adjusted Wilcoxon signed-rank $p$-values. Table~\ref{tab:prompt_tests} adds bootstrap confidence intervals, permutation tests and Benjamini--Hochberg correction.

\item {\bf Experiments compute resources}
    \item[] Question: For each experiment, does the paper provide sufficient information on the computer resources (type of compute workers, memory, time of execution) needed to reproduce the experiments?
    \item[] Answer: \answerYes{} 
    \item[] Justification: Appendix A (Compute paragraph and Table 2) gives GPU types, per-model GPU counts and the total GPU-hours.
    
\item {\bf Code of ethics}
    \item[] Question: Does the research conducted in the paper conform, in every respect, with the NeurIPS Code of Ethics \url{https://neurips.cc/public/EthicsGuidelines}?
    \item[] Answer: \answerYes{} 
    \item[] Justification: We have read and meet the NeurIPS code of ethics.

\item {\bf Broader impacts}
    \item[] Question: Does the paper discuss both potential positive societal impacts and negative societal impacts of the work performed?
    \item[] Answer: \answerYes{} 
    \item[] Justification: We discuss this in the conclusion.
    
\item {\bf Safeguards}
    \item[] Question: Does the paper describe safeguards that have been put in place for responsible release of data or models that have a high risk for misuse (e.g., pre-trained language models, image generators, or scraped datasets)?
    \item[] Answer: \answerNA{} 
    \item[] Justification: We do not release any new data or models.

\item {\bf Licenses for existing assets}
    \item[] Question: Are the creators or original owners of assets (e.g., code, data, models), used in the paper, properly credited and are the license and terms of use explicitly mentioned and properly respected?
    \item[] Answer: \answerYes{} 
    \item[] Justification: The dataset we use is covered under an Apache 2.0 license, and we properly credit and cite their work.

\item {\bf New assets}
    \item[] Question: Are new assets introduced in the paper well documented and is the documentation provided alongside the assets?
    \item[] Answer: \answerNA{} 
    \item[] Justification: We do not release new assets.

\item {\bf Crowdsourcing and research with human subjects}
    \item[] Question: For crowdsourcing experiments and research with human subjects, does the paper include the full text of instructions given to participants and screenshots, if applicable, as well as details about compensation (if any)? 
    \item[] Answer: \answerNA{} 
    \item[] Justification: We do not personally do any crowdsourced research. We build upon the open assistant dataset, which contains human responses.

\item {\bf Institutional review board (IRB) approvals or equivalent for research with human subjects}
    \item[] Question: Does the paper describe potential risks incurred by study participants, whether such risks were disclosed to the subjects, and whether Institutional Review Board (IRB) approvals (or an equivalent approval/review based on the requirements of your country or institution) were obtained?
    \item[] Answer: \answerNA{} 
    \item[] Justification: Our research did not involve human subjects.

\item {\bf Declaration of LLM usage}
    \item[] Question: Does the paper describe the usage of LLMs if it is an important, original, or non-standard component of the core methods in this research? Note that if the LLM is used only for writing, editing, or formatting purposes and does \emph{not} impact the core methodology, scientific rigor, or originality of the research, declaration is not required.
    \item[] Answer: \answerNA{} 
    \item[] Justification: We do not use LLMs for any important, original, or non-standard components, only as a programming aid, as mentioned by the NeurIPS handbook.

\end{enumerate}

\end{document}